\documentclass[lettersize,journal]{IEEEtran}
\usepackage{amsmath,amsfonts}
\usepackage{algorithm}
\usepackage{array}
\usepackage[caption=false,font=normalsize,labelfont=sf,textfont=sf]{subfig}
\usepackage{textcomp}
\usepackage{stfloats}
\usepackage{url}
\usepackage{verbatim}
\usepackage{graphicx}
\usepackage{cite}
\usepackage{placeins}

\IEEEoverridecommandlockouts

\usepackage{algpseudocode} 
\usepackage{float}
\usepackage{subcaption}
\usepackage{multirow}
\usepackage{booktabs}
\usepackage{array}

\begin{document}

\title{Toward Accountable AI-Generated Content on Social Platforms: Steganographic Attribution and Multimodal Harm Detection}
\author{
    Xinlei Guan,
    David Arosemena,
    Tejaswi Dhandu,
    Kuan Huang,
    Meng Xu,
    Miles Q. Li,
    Bingyu Shen,

    Ruiyang Qin,
    Umamaheswara Rao Tida,
    Boyang Li

\thanks{X. Guan, D. Arosemena, K. Huang, M. Xu, and B. Li are with Kean University, Union, NJ 07083 USA (e-mail: \{guanxi, arosemed, kuan.huang, meng.xu, boli\}@kean.edu).}%
\thanks{T. Dhandu and U. R. Tida are with North Dakota State University, Fargo, ND 58102 USA (e-mail: \{tejaswi.dhandu, umamaheswara.tida\}@ndsu.edu).}%
\thanks{M. Q. Li is with McGill University, Montréal, QC H3A 0G4, Canada (e-mail: miles.qi.li@mail.mcgill.ca).}%
\thanks{R. Qin is with Villanova University, Villanova, PA 19085 USA (e-mail: rqin@villanova.edu).}%
\thanks{B. Shen is with the University of Notre Dame, Notre Dame, IN 46556 USA (e-mail: bingyu.shen@hotmail.com).}%
\thanks{Corresponding author: Boyang Li (e-mail: boli@kean.edu).}%
\thanks{This work has been submitted to the IEEE for possible publication. Copyright may be transferred without notice, after which this version may no longer be accessible.}%
}



\IEEEpubid{0000--0000/00\$00.00~\copyright~2021 IEEE}

\maketitle
\IEEEpubid{}

\begin{abstract}
The rapid growth of generative AI has introduced new challenges in content moderation and digital forensics. In particular, benign AI-generated images can be paired with harmful or misleading text, creating difficult-to-detect misuse. This contextual misuse undermines the traditional moderation framework and complicates attribution, as synthetic images typically lack persistent metadata or device signatures. We introduce a steganography enabled attribution framework that embeds cryptographically signed identifiers into images at creation time and uses multimodal harmful content detection as a trigger for attribution verification. Our system evaluates five watermarking methods across spatial, frequency, and wavelet domains. It also integrates a CLIP-based fusion model for multimodal harmful-content detection. Experiments demonstrate that spread-spectrum watermarking, especially in the wavelet domain, provides strong robustness under blur distortions, and our multimodal fusion detector achieves an AUC-ROC of 0.99, enabling reliable cross-modal attribution verification. These components form an end-to-end forensic pipeline that enables reliable tracing of harmful deployments of AI-generated imagery, supporting accountability in modern synthetic media environments. Our code is available at GitHub: https://github.com/bli1/steganography
\end{abstract}


\section{Introduction}
\label{sec:intro}

The proliferation of generative AI tools has fundamentally altered the landscape of digital content creation~\cite{goodfellow2014generative, ho2020denoising, ramesh2021zero}. Users can now generate photorealistic images with minimal effort, which creates significant challenges for content moderation and digital forensics. While existing approaches to harmful content detection typically evaluate images in isolation, real-world misuse often depends on context: a generated image that violates no policies on its own may become a vehicle for harmful content or hate speech when paired with misleading captions or posted in inflammatory contexts on social platforms. AI-generated images become harmful when combined with context.~\cite{kiela2020hateful} This highlights a key limitation of current moderation systems: they fail to capture harm that emerges from multimodal context.

Consider a scenario in which a user generates a realistic image depicting a crowded street scene in a foreign city. The image itself is innocuous, with no violence, no identifiable individuals, and no policy violations. However, when posted to social media with a caption such as “I know your address” or other threatening messages, the combination becomes harmful and intimidating. The generated image lends visual credibility to the threat, even though the image itself contains no sensitive information. Traditional content moderation frameworks evaluating the image alone would find nothing objectionable. Benign images can become harmful through misleading or malicious captions.~\cite{hee2024recent} To address this limitation, we design an end-to-end forensic framework for attribution.

\begin{figure}[tb]
\centering
\includegraphics[width=0.9\linewidth, trim=0 0 0 0, clip]{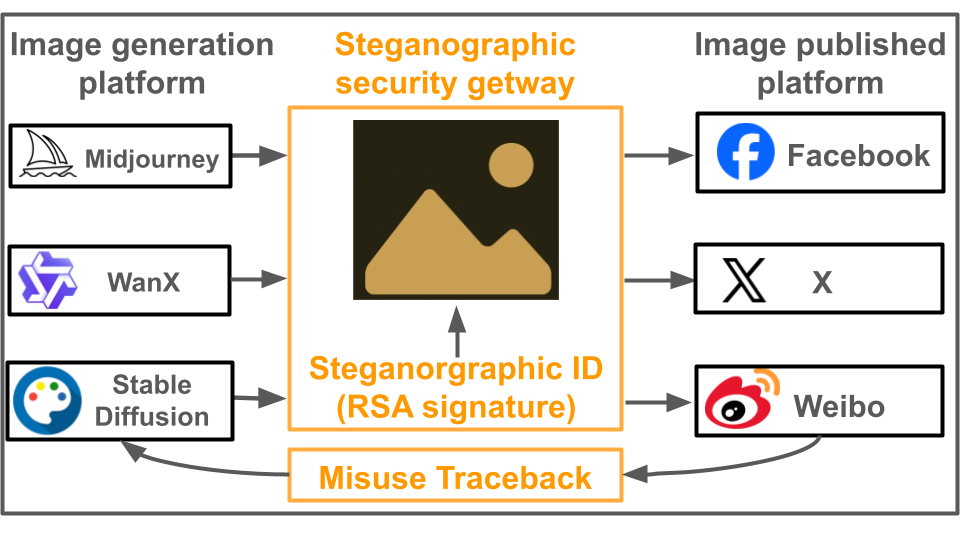}
\caption{Overview of the Steganographic Security Gateway.}
\label{fig:img_6}
\end{figure}

Consequently, this contextual deficiency poses a significant forensic hurdle with practical implications, as illustrated in Figure~\ref{fig:img_6}, which outlines the suggested framework for its resolution. Synthetic media manipulations have been implicated in election interference, stock market manipulation, and targeted harassment initiatives~\cite{kaushik2025financial, de2024beyond}. Furthermore, the increasing accessibility of generative tools has substantially lowered the threshold for creating persuasive visual content,
thereby empowering malicious entities to generate deceptive imagery on a large scale \cite{brundage2018malicious}. The lack of attribution makes misuse hard to investigate. Platforms struggle to respond effectively, as detection and removal typically lag behind content propagation, and the original poster may have deleted their account or obscured their identity. As a result, investigators are left with limited recourse \cite{diresta2024aiimages}.

The image itself carries no inherent link to its creator, and metadata can be easily stripped or falsified. Unlike traditional photography, where device fingerprints or EXIF data may offer attribution clues, AI-generated images emerge from cloud-based services that often retain minimal logs or lack mechanisms for post-hoc attribution. AI images lack metadata for attribution. This attribution vacuum undermines accountability and emboldens misuse. Research on security in deep learning has progressed from adversarial attacks~\cite{goodfellow2014generative, goodfellow2014explaining, gong2019real} on neural networks to more recent concerns such as hallucination and safety failures in vision-language models~\cite{ye2025survey, Hong_2026_WACV}. As generative models become more widely available, attribution becomes increasingly important, as it offers the possibility of deterrence that may reduce misuse in the first place.

We propose an accountability framework that embeds cryptographic identifiers into AI-generated images at the point of creation using steganographic techniques, enabling forensic tracing from detected misuse back to the originating user. Figure~\ref{fig:img_6} illustrates the overall framework, where an AI-generated image is first embedded with a cryptographic identifier, then distributed through platforms, and finally analyzed for harmful content before attribution is performed. Our framework operates in two stages. First, a multimodal classifier analyzes image–text pairs to detect harmful content. Second, for content flagged as harmful, the framework performs attribution by decoding and validating embedded cryptographic identifiers, enabling attribution to the originating user. In practice, this framework is intended to be integrated with generative models, where identity information is embedded at the time of image generation. Our approach is model-agnostic and evaluates multiple information hiding methods spanning spatial, frequency, and wavelet domains to assess robustness across realistic platform conditions. 

The primary contribution of this work is the end-to-end forensic pipeline rather than the embedding techniques themselves: from identifying contextually harmful content, to extracting embedded ciphertext via the appropriate steganographic decoder, resolving user identity through a controlled decryption process. This framework offers investigators a practical tool for attribution in cases where generative AI outputs are weaponized through contextual manipulation, thereby supporting accountability in an increasingly synthetic media environment.

\section{Background}

\subsection{Generative AI and Synthetic Media}
Recent advances in deep generative models have transformed synthetic media from a research curiosity into a mainstream capability. Diffusion models, which generate images through iterative denoising processes, have largely supplanted earlier generative adversarial network (GAN) approaches due to their superior image quality and training stability. Models such as Stable Diffusion\cite{rombach2022high}, DALL-E\cite{ramesh2021zero}, and Midjourney\cite{oppenlaender2022creativity} can produce photorealistic images from natural language prompts, enabling users with no artistic or technical expertise to create convincing visual content in seconds.

The accessibility of these tools has driven rapid adoption. Commercial platforms offer generation capabilities through easily accessible interfaces, while open-source models allow unrestricted local deployment. This democratization led to an explosion in synthetic content volume: millions of AI-generated images are created daily across various platforms. While most of them serve legitimate creative, commercial, and educational purposes, the same accessibility that empowers benign users also lowers barriers for those with malicious intent. The combination of high output quality and low production cost creates an environment where synthetic media can be manufactured at scale for harmful content, fraud, or harassment campaigns \cite{loth2024blessing, shoaib2023deepfakes}.

\subsection{Content Moderation and Its Limitations}
Social media platforms have developed increasingly sophisticated content moderation frameworks to address harmful material\cite{gongane2022detection}. These frameworks typically employ a combination of automated classifiers, perceptual hash matching against known harmful content databases, and human review for flagged or ambiguous cases. Image-based classifiers are trained to detect specific categories of policy-violating content such as violence, nudity, or extremist imagery, enabling rapid screening of uploaded material.

However, these approaches operate primarily on images in isolation, evaluating visual content independent of the context in which it appears. This paradigm fails to capture cases where harm emerges from the interaction between an image and its surrounding context\cite{yuan2024rethinking, aneja2023cosmos}. A generated image of a protest scene may be entirely benign when posted with accurate information, yet become harmful content when paired with a fabricated caption attributing it to a different event or location. Similarly, an innocuous portrait could become a vehicle for harassment when posted with false claims about the depicted individual.

The challenge is compounded by the multimodal nature of social media content. Posts typically combine images with text, and the semantic relationship between these modalities can be difficult to assess. Current multimodal classifiers have made progress in joint image-text understanding\cite{kiela2020hateful}, but detecting contextual manipulation—where individually acceptable components combine to produce harmful messaging—remains an open problem\cite{gomez2020exploring}. Content moderation operates reactively, addressing harmful posts only after reporting, often following significant spread—limiting deterrence as bad actors face minimal consequences even upon removal.

\subsection{Steganography and Information Hiding}
Steganography, derived from the Greek words for ``covered writing", is the practice of concealing secret messages within innocuous carrier media such that the existence of the hidden communication is not apparent to observers. While watermarking and steganography share technical foundations—both involve embedding information into media—they differ fundamentally in purpose and design priorities. Watermarking prioritizes robustness, seeking to create marks that persist even when an adversary attempts to remove them. Steganography prioritizes undetectability, seeking to hide the very existence of the embedded message from adversarial scrutiny.

Classical steganographic techniques for AI-generated images include least significant bit (LSB) \cite{van1994digital} embedding, where message bits replace the lowest-order bits of pixel values, and transform domain methods that modify discrete cosine transform (DCT) \cite{cox1997secure} or wavelet coefficients\cite{xia1997multiresolution}. More recent approaches leverage deep learning, training encoder networks to embed messages in ways that minimize statistical detectability while decoder networks recover the hidden content. The capacity-detectability tradeoff is central to steganographic framework design: embedding more information generally increases the statistical footprint that steganalysis techniques can exploit.

For the accountability framework proposed in this work, steganographic principles inform the design of the embedding mechanism. The cryptographic payload identifying the originating user must be embedded in a manner that survives the AI-generated image's lifecycle on social platforms—including compression, resizing, and format conversion—while remaining imperceptible to users. This requirement blends watermarking's robustness goals with steganography's concealment objectives. The embedded ciphertext must not be trivially detectable or removable by malicious users seeking to evade attribution, yet must remain extractable by authorized investigators examining flagged content.

\subsection{Digital Forensics for Synthetic Media}
The digital forensics community has developed various techniques for detecting AI-generated content\cite{verdoliva2020media, wang2020cnn, rossler2019faceforensics}. 
However, detection alone addresses only part of the forensic challenge. Determining that an image is AI-generated provides no information about who created it, when, using which tool, or for what purpose. This attribution gap is problematic in investigative contexts. When a synthetic image surfaces as part of a disinformation campaign or harassment effort, investigators need to trace responsibility to hold bad actors accountable and potentially identify broader coordinated operations. Current synthetic media detectors offer a binary classification—real or fake—without the provenance information necessary for attribution.

Traditional digital forensics relies on metadata and device fingerprints to establish attribution. Photographs captured by physical cameras contain EXIF data recording device information, timestamps, and sometimes GPS coordinates, while sensor imperfections create device-specific fingerprints that can link images to particular cameras. AI-generated images lack these physical-world anchors. They are produced by software processes that typically do not embed meaningful metadata, and cloud-based generation services may strip or anonymize any information that could identify users. This absence of inherent provenance information creates an accountability vacuum that the framework proposed in this work aims to address.

\subsection{Provenance and Accountability frameworks}
Recognizing the need for content authenticity verification, several industry initiatives have emerged to establish provenance standards for digital media. The Content Authenticity Initiative (CAI) and Coalition for Content Provenance and Authenticity (C2PA)\cite{rosenthol2022c2pa} have developed specifications for cryptographically signed metadata that travels with media files, recording information about origin, creation tools, and subsequent edits. These standards enable consumers and platforms to verify claims about content provenance by checking cryptographic signatures against trusted certificates.
While these initiatives represent meaningful progress, they face limitations in adversarial contexts. C2PA metadata is stored alongside or within media files and can be stripped by users seeking to obscure provenance. Platforms that re-encode uploaded content may not preserve provenance metadata, breaking the chain of authenticity. Furthermore, these frameworks focus primarily on establishing that content is authentic rather than enabling post-hoc attribution of misuse. They answer the question ``was this image really produced by the claimed tool?" rather than ``who is responsible for this harmful deployment?"
Platform-level logging offers another potential avenue for accountability. Generation services could maintain records linking created content to user accounts, enabling retrospective identification when misuse is reported. However, this approach raises privacy concerns, requires cooperation from generation platforms that may operate across jurisdictions, and fails when users employ open-source tools locally or when content crosses platform boundaries. The gap between the existing provenance framework and the attribution needs of forensic investigators motivates the framework developed in this work: a mechanism for embedding user-identifying information directly into generated content in a manner that enables controlled tracing when contextual misuse is detected.

\section{Method}
At the beginning of this section, we define the notation used throughout the pipeline. We use $\tau_{\text{cls}}$ to denote the classification threshold and $\tau_{\text{corr}}$ to denote the correlation threshold used in watermark verification. In this work, we focus on watermarking for attribution, where the primary objective is robustness to common image transformations rather than strict undetectability. While the embedding process shares characteristics with steganographic techniques, our design prioritizes reliable recovery of embedded identity information under real-world distortions.
\label{sec:method}

\begin{figure}[t]
  \centering
  \fbox{%
    \includegraphics[width=0.9\linewidth]{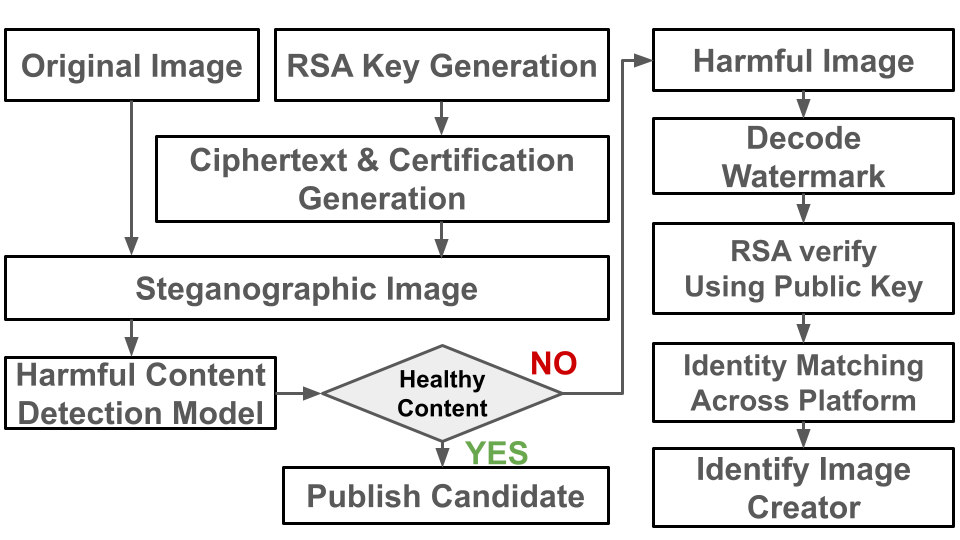}
  }
  \caption{Flowchart of the Steganographic Enabled Image Tracing Pipeline.}
  \label{fig:flowchat}
\end{figure}

\subsection{Dataset}
\label{sec:dataset}





The framework takes as input image–text pairs (I, T). The proposed framework assumes that images are generated by AI models, where a cryptographic identifier can be embedded at the time of creation.  For evaluation, we use the multimodal safety dataset from~\cite{wang-etal-2025-cant}, which is designed to study how safety issues can arise when images and text are interpreted together rather than separately. The dataset contains paired image and text samples with safety labels, including both benign and harmful cases, and many examples are harmless on their own but become harmful content when combined with certain captions. We use the term "harmful content" to refer to image–text pairs that convey misleading, abusive, or disinformation-based intent. In our experiments, we use the dataset as it is, without any changes, to evaluate our watermark-based attribution capability method in realistic multimodal settings. While the dataset provides real-world multimodal examples, it serves as a proxy for AI-generated content in our evaluation.

\subsection{Steganography Algorithm}

Let $I$ denote the original AI-generated image, $I'$ the encoded (watermarked) image, and $I''$ the post-distribution image after platform transformations. We evaluate five watermarking techniques that operate in distinct embedding domains: spatial LSB embedding, DCT-domain embedding, DWT-domain embedding, spatial spread spectrum watermarking, and wavelet domain spread spectrum watermarking (DWT SS). Figure~\ref{fig:flowchat} illustrates the complete pipeline, namely embedding, platform transformation, detection, and attribution stages. The objective of the embedding process is to maximize payload recoverability under realistic image transformations while maintaining imperceptible visual distortion. After embedding, the watermarked image may be distributed through online platforms, where it can transform such as compression, resizing, or filtering. We model this stage as a transition from the encoded image $I'$(watermarked image) to a potentially distorted image $I''$(post-distribution image), which serves as the input to the verification stage. Formally, the pipeline takes as input an image–text pair (I, T) and produces a harmfulness decision $\hat{y}$. If $\hat{y}$ = 1, the framework triggers the attribution, where the embedded identifier is decoded and verified to recover the source identity. Thus, harmful content detection acts as a gating mechanism for attribution.

Although these methods differ in how their image coefficients are modified, they share the common goal of embedding a hidden identifier without introducing perceptible visual distortion.
The three bit-level embedding schemes (LSB, DCT, and DWT) embed bits directly into the spatial domain or transform-domain coefficients of the host image, following the per-image processing procedure described in Algorithm~\ref{alg:process-one}. LSB substitutes the least significant bits of blue channel samples, and this technique offers high capacity but low robustness. The DCT method embeds watermark bits within 8×8 blocks by adjusting selected mid-frequency coefficients to satisfy parity constraints. The DWT method embeds bits in the LH subband of a Haar wavelet transform by applying parity control to quantized coefficients.

The spread spectrum techniques rely on noise modulation rather than direct bit replacement. Spatial spread spectrum divides the flattened image into 32 segments and adds $\pm \alpha$ pseudo-random noise according to each watermark bit. The parameter $\alpha$ denotes the embedding strength that controls the magnitude of the perturbation. Detection is performed through correlation and bit error rate analysis. The DWT-SS method applies a similar modulation strategy to the LL subband of a wavelet decomposition, which provides increased resilience to low-pass distortions such as blur.

These individual techniques collectively allow a comprehensive comparison across spatial, frequency, and wavelet-domain watermarking paradigms, with unified encoding and verification steps formalized in Algorithm~\ref{alg:process-one}. In Algorithm 1, $W$ denotes the plaintext identifier associated with the signed payload and is used only for method-specific verification and logging.

\begin{algorithm}[t]
  \caption{Processing a Single Image with Five watermark Schemes}
  \label{alg:process-one}
  \begin{algorithmic}[1]
    \Require Image $I$, RSA signature bytes $S$, bit-length $L$, watermark text $W$
    \Ensure Validity flags $(v_{\text{LSB}}, v_{\text{DCT}}, v_{\text{DWT}}, v_{\text{SS}}, v_{\text{DWT-SS}})$

    \State Create subfolders for $I$: encoded, decoded, comparison
    \State Convert $I$ to 3-channel format if necessary

    \Statex \Comment{--- LSB ---}
    \State $I_{\text{LSB}} \gets \textsc{EncodeLSB}(I, S)$
    \State $v_{\text{LSB}} \gets \textsc{VerifyPath}(I_{\text{LSB}}, \text{LSB}, S, L, W)$

    \Statex \Comment{--- DCT ---}
    \State $I_{\text{DCT}} \gets \textsc{EncodeDCT}(I, S)$
    \State $v_{\text{DCT}} \gets \textsc{VerifyPath}(I_{\text{DCT}}, \text{DCT}, S, L, W)$

    \Statex \Comment{--- DWT ---}
    \State $I_{\text{DWT}} \gets \textsc{EncodeDWT}(I, S)$
    \State $v_{\text{DWT}} \gets \textsc{VerifyPath}(I_{\text{DWT}}, \text{DWT}, S, L, W)$

    \Statex \Comment{--- Spatial Spread Spectrum (SS) ---}
    \State $I_{\text{SS}} \gets \textsc{EncodeSS}(I, S)$
    \State $v_{\text{SS}} \gets \textsc{VerifyPath}(I_{\text{SS}}, \text{SS}, S, L, W)$

    \Statex \Comment{--- DWT-domain Spread Spectrum (DWT-SS) ---}
    \State $I_{\text{DWT-SS}} \gets \textsc{EncodeDWTSS}(I, S)$
    \State $v_{\text{DWT-SS}} \gets \textsc{VerifyPath}(I_{\text{DWT-SS}}, \text{DWT-SS}, S, L, W)$

    \State Save $(I\_name, v_{\text{LSB}}, v_{\text{DCT}}, v_{\text{DWT}},
           v_{\text{SS}}, v_{\text{DWT-SS}})$ into the Excel row

    \State \Return $(v_{\text{LSB}}, v_{\text{DCT}}, v_{\text{DWT}}, v_{\text{SS}}, v_{\text{DWT-SS}})$
  \end{algorithmic}
\end{algorithm}

\subsection{Asymmetric Encryption}

Decoding and verification are performed only for images flagged as harmful ($\hat{y} = 1$) by the multimodal classification stage. This decoding process is enabled by an asymmetric encryption framework, which unifies all watermark methods and provides a shared cryptographic identity layer across heterogeneous embedding schemes, as formalized in Algorithm~\ref{alg:unified}. The watermark payload is generated based on the predefined encoding rules. Here, R denotes the predefined payload construction rules that specify how user identity information is serialized into a signable message format before embedding. It is signed using a 1024-bit RSA private key. The resulting digital signature ensures authenticity, prevents forgery, and establishes a cryptographically verifiable attribution.

For LSB, DCT, and the DWT watermark, the RSA signature is transformed into a binary bitstream that is directly embedded into the image. During extraction, the recovered bitstream is verified using the corresponding RSA public key. For the spread-spectrum schemes, we use a more compact 32-bit fingerprint derived from the SHA-256 hash of the RSA signature. Different payload representations are used to match the characteristics of each embedding scheme. Bit-level methods embed the full RSA signature to enable exact verification, while spread-spectrum methods use a compact fingerprint derived from the signature to support correlation-based detection under noisy conditions. This fingerprint maintains the asymmetric verification property while improving the stability of correlation-based detection.

\begin{algorithm}[t]
  \caption{Unified Asymmetric-Cryptographic watermark Framework}
  \label{alg:unified}
  \begin{algorithmic}[1]
    \Require Image $I$, user identifier $u$, payload rules $\mathcal{R}$
    \Ensure Validity flags $v_{\mathsf{m}}$ for all watermark schemes

    \State \textbf{Asymmetric payload generation}
    \State $P \gets \textsc{GeneratePayload}(u, \mathcal{R})$
    \If{RSA key pair $(sk, pk)$ not found}
      \State generate 1024-bit RSA key pair $(sk, pk)$
    \EndIf
    \State $S \gets \text{Sign}_{sk}(P)$; \quad $L_S \gets \text{bit\_length}(S)$
    \State $F \gets \textsc{TruncateTo32Bits}(\text{SHA256}(S))$

    \Statex
    \State \textbf{Embedding phase}
    \State Convert $I$ to 3-channel format if needed
    \For{each bit-embedding scheme $\mathsf{m} \in \{\text{LSB},\text{DCT},\text{DWT}\}$}
      \State $I_{\mathsf{m}} \gets \textsc{EncodeBitScheme}(I, S, \mathsf{m})$
    \EndFor
    \For{each spread-spectrum scheme $\mathsf{m} \in \{\text{SS},\text{DWT-SS}\}$}
      \State $I_{\mathsf{m}} \gets \textsc{EncodeSpreadSpectrum}(I, F, \mathsf{m})$
    \EndFor

    \Statex
    \State \textbf{Verification phase}
    \For{each bit-embedding scheme $\mathsf{m} \in \{\text{LSB},\text{DCT},\text{DWT}\}$}
      \State $S' \gets \textsc{DecodeBits}(I_{\mathsf{m}}, L_S, \mathsf{m})$
      \State $v_{\mathsf{m}} \gets \text{Verify}_{pk}(P, S')$
    \EndFor
    \For{each spread-spectrum scheme $\mathsf{m} \in \{\text{SS},\text{DWT-SS}\}$}
      \State $F' \gets \textsc{DetectFingerprint}(I_{\mathsf{m}}, \mathsf{m})$
      \State $v_{\mathsf{m}} \gets (\text{Corr}(F', F) \ge \tau)$
    \EndFor

    \State \Return $\{v_{\mathsf{m}}\}$ for all schemes
  \end{algorithmic}
\end{algorithm}

If verification succeeds (i.e., $v_m = 1$), the embedded identity is considered authentic and untampered, and can be used for attribution. 

Overall, the asymmetric-encryption layer provides a secure and verifiable foundation across all watermark approaches, ensuring that the embedded identity token cannot be fabricated or altered without possession of the private key, as enforced by the end-to-end signing, embedding, and verification pipeline in Algorithm~\ref{alg:unified}.


\subsection{VLM-Based Multimodal Encoding}
\label{subsec:vlm-encoding}

We adopt a pre-trained CLIP model as a frozen vision--language encoder. Given an input  AI-generated image $I$ and the associated text $T$, we denote the image encoder by $f_{\text{img}}(\cdot)$ and the text encoder by $f_{\text{txt}}(\cdot)$. The encoders map both modalities into a shared $d$-dimensional embedding space:
\begin{equation}
    \mathbf{e}_{\text{img}} = f_{\text{img}}(I) \in \mathbb{R}^d, 
    \quad
    \mathbf{e}_{\text{txt}} = f_{\text{txt}}(T) \in \mathbb{R}^d.
\end{equation}

Following the standard CLIP practice, we apply $\ell_2$ normalization to both embeddings:
\begin{equation}
    \tilde{\mathbf{e}}_{\text{img}} = 
    \frac{\mathbf{e}_{\text{img}}}{\|\mathbf{e}_{\text{img}}\|_2},
    \qquad
    \tilde{\mathbf{e}}_{\text{txt}} = 
    \frac{\mathbf{e}_{\text{txt}}}{\|\mathbf{e}_{\text{txt}}\|_2}.
\end{equation}

The normalized vectors $\tilde{\mathbf{e}}_{\text{img}}$ and $\tilde{\mathbf{e}}_{\text{txt}}$ capture high-level semantics of the visual and textual content in a compatible feature space. Throughout training and inference, the CLIP backbone is kept frozen; only the subsequent fusion classifier is trained, which makes our detector lightweight and easy to deploy. The embedding dimension d is determined by the selected pre-trained CLIP backbone.

\subsection{Multimodal Fusion Representation}
\label{subsec:multimodal-fusion}

A key design choice of our detector is the fusion representation that combines the two modality-specific embeddings into a single feature vector. Instead of relying solely on cosine similarity or simple concatenation, we explicitly model alignment, discrepancy, and interaction between the image and text embeddings.

Given the normalized embeddings $\tilde{\mathbf{e}}_{\text{img}}$ and $\tilde{\mathbf{e}}_{\text{txt}}$, we construct four vector features and one scalar:

\textbf{Raw embeddings.}
We first keep the original  AI-generated image and text embeddings:
\begin{equation}
    \mathbf{z}_{\text{img}} = \tilde{\mathbf{e}}_{\text{img}},
    \qquad
    \mathbf{z}_{\text{txt}} = \tilde{\mathbf{e}}_{\text{txt}}.
\end{equation}

\textbf{Difference (discrepancy).}
To capture the semantic mismatch between the visual and textual content, we compute the element-wise difference:
\begin{equation}
    \mathbf{z}_{\text{diff}} = \tilde{\mathbf{e}}_{\text{img}} - \tilde{\mathbf{e}}_{\text{txt}}.
\end{equation}
Large values in $\mathbf{z}_{\text{diff}}$ indicate dimensions where the two modalities disagree, which can be informative for detecting inconsistent or misleading descriptions.

\textbf{Element-wise product (interaction).}
To model cross-modal interactions, we compute the element-wise product:
\begin{equation}
    \mathbf{z}_{\text{prod}} = \tilde{\mathbf{e}}_{\text{img}} \odot \tilde{\mathbf{e}}_{\text{txt}},
\end{equation}
where $\odot$ denotes the Hadamard product. This term highlights dimensions that are simultaneously activated in both embeddings, approximating joint support over semantic concepts.

\textbf{Cosine similarity (global alignment).}
We also include a scalar cosine similarity:
\begin{equation}
    s = \cos\bigl(\tilde{\mathbf{e}}_{\text{img}}, \tilde{\mathbf{e}}_{\text{txt}}\bigr)
      = \tilde{\mathbf{e}}_{\text{img}}^{\top} \tilde{\mathbf{e}}_{\text{txt}} \in \mathbb{R},
\end{equation}
which summarizes the overall alignment between the image and text in the CLIP embedding space.

\textbf{Fused feature vector.}
Finally, we concatenate all components into a single fusion vector:
\begin{equation}
    \mathbf{z} = 
    \bigl[
        \mathbf{z}_{\text{img}} \,;\,
        \mathbf{z}_{\text{txt}} \,;\,
        \mathbf{z}_{\text{diff}} \,;\,
        \mathbf{z}_{\text{prod}} \,;\,
        s
    \bigr]
    \in \mathbb{R}^{4d + 1},
\end{equation}
where $[\,\cdot\,;\,\cdot\,]$ denotes vector concatenation. This fused representation jointly encodes (1) modality-specific semantics, (2) discrepancies between modalities, (3) multiplicative interactions, and (4) a global similarity score, providing a rich feature basis for harmful content detection.

\subsection{Harmful Content Classification}
\label{subsec:harmful-classifier}

The fused vector $\mathbf{z}$ is passed to a lightweight multilayer perceptron (MLP) classifier that outputs the probability that the image-text pair is harmful. Concretely, we use a two-layer MLP with ReLU activations and dropout:
\begin{align}
    \mathbf{h}_1 &= \phi\bigl( \mathbf{W}_1 \mathbf{z} + \mathbf{b}_1 \bigr), \\
    \mathbf{h}_2 &= \phi\bigl( \mathbf{W}_2 \mathbf{h}_1 + \mathbf{b}_2 \bigr), \\
    o &= \mathbf{w}_3^{\top} \mathbf{h}_2 + b_3,
\end{align}
where $\phi(\cdot)$ denotes the element-wise ReLU nonlinearity, and dropout is applied after each hidden layer during training. The scalar output $o$ is converted into a harmfulness probability via the sigmoid function:
\begin{equation}
    p = \sigma(o) = \frac{1}{1 + e^{-o}}.
\end{equation}

Given a decision threshold $\tau \in (0,1)$, the predicted label $\hat{y}$ is:
\begin{equation}
    \hat{y} =
    \begin{cases}
        1, & \text{if } p \ge \tau \quad (\text{harmful}), \\
        0, & \text{if } p < \tau \quad (\text{benign}).
    \end{cases}
\end{equation}

The output of this stage is a harmfulness label $\hat{y} \in \{0,1\}$. If $\hat{y} = 1$, the image–text pair is forwarded to the attribution stage for decoding and identity recovery; otherwise, no attribution step is performed. This decision flow is consistent with the pipeline shown in Figure~\ref{fig:flowchat}.

During training, we minimize the binary cross-entropy loss over a dataset of labeled image--text pairs $\{(I_i, T_i, y_i)\}_{i=1}^N$:
\begin{equation}
    \mathcal{L} = - \frac{1}{N} \sum_{i=1}^{N}
    \bigl[
        y_i \log p_i + (1 - y_i) \log (1 - p_i)
    \bigr],
\end{equation}

where $p_i$ is the predicted harmfulness probability for the $i$-th pair. Only the parameters of the MLP classifier are updated; the CLIP backbone remains frozen. This design keeps the detector compact and training-efficient while still exploiting the strong multimodal representations learned by large VLMs. It is important to note that the model performs statistical classification based on learned multimodal representations rather than explicit reasoning about intent or context.

\subsection{Identity Matching on Third-Party Platforms}
\label{subsec:identity-matching}

The input to this stage is a verified identity token extracted from the  AI-generated image, along with authorized platform metadata when available. This stage assumes that authorized investigators have access to sufficient platform metadata or cross-platform signals to map the recovered identity token to a user. The output is a resolved user identity or a set of candidate matches associated with the content. This stage operates outside the automated pipeline and depends on access to external platform data under appropriate authorization. As such, it is not evaluated as part of the core framework, but serves as a downstream forensic step enabled by the verified identity token. 

Under permitted conditions, investigators may correlate accounts across different platforms using available metadata and behavioral signals. Common linkage signals include consistent usernames, profile attributes, activity patterns, and device- or upload-related artifacts. When authorized, browser-level data such as \emph{third-party cookies} and related storage artifacts may also provide auxiliary evidence for cross-platform identity correlation.

Identity matching complements the attribution pipeline: After harmful content is identified and its embedded provenance data is decoded, investigators may combine the recovered identity information with platform metadata to contextualize the origin and distribution of the content.

Overall, the framework operates as a two-stage pipeline. Given an image–text pair $(I, T)$, the framework first performs harmful-content classification to produce a label $\hat{y}$. If $\hat{y} = 1$, the framework proceeds to decode and verify the embedded identity, and, when possible, resolves the source user through authorized matching.

The final output of the framework consists of a harmfulness decision $\hat{y}$ and, when attribution is successful, a corresponding source identity. This process can be summarized as a mapping $(I, T) \rightarrow \hat{y} \rightarrow \text{identity}$.

\section{Experiment}

Experimentally, we evaluate individual components of the pipeline, including watermark robustness and multimodal harmful content detection, which together support the overall attribution workflow. This section presents the experimental setup, qualitative observations, and quantitative results for all five watermarking techniques (LSB, DCT, DWT), spatial spread spectrum, and wavelet domain spread spectrum (DWT-SS). To ensure meaningful comparison, every individual technique was implemented within a unified local testing framework, sharing identical directory structures, RSA key pairs, decoding routines, and logging procedures.

\subsection{Dataset and Experimental Setup}

All experiments were conducted on a diverse collection of PNG, JPEG, and BMP images placed under a common \texttt{Original\_image} directory. This dataset intentionally mixed different compression levels and color depths to approximate real-world image variability rather than relying on idealized conditions.  

For the bit-level embedding individual techniques (LSB, DCT, DWT), the pipeline automatically generated a dedicated subdirectory for each source image. Watermarked outputs were saved as \texttt{LSB.png}, \texttt{DCT.png}, and \texttt{DWT.png} in \texttt{Encoded\_image}, while corresponding decoded RSA signatures were stored under \texttt{Decoded\_output}. An Excel summary was generated for each image to record verification outcomes.

The spread spectrum schemes required additional storage paths to preserve all stages of processing. For both spatial domain SS and DWT-SS, embedded images, decoded correlation values, and bit error rate (BER) statistics were placed in method-specific folders such as \texttt{Spatial\_encoded}, \texttt{Spatial\_decoded}, and \texttt{Spread\_comparison}. Both individual techniques used a 32-bit fingerprint derived from the RSA signature, enabling correlation-based detection even when numerical distortions occurred.

A final stage introduced robustness testing. All previously watermarked images were subjected to realistic post-processing distortions after embedding. In addition to Gaussian blur (radius = 0.5), we further applied JPEG compression with quality factors Q=50, as well as a geometric transformation consisting of resizing the image to 80\% of its original resolution followed by center cropping back to the original size. The distorted counterparts were saved in parallel folder structures, and their decoding and verification results were recorded in the same manner as the pristine images. This allowed a direct comparison between clean and attacked conditions across multiple transformation scenarios, supporting a controlled and reproducible evaluation of robustness.

\subsection{Qualitative Analysis of Embedding Artifacts}

Before analyzing the numerical results, we examined the visual characteristics produced by each embedding strategy. These qualitative observations provided insight into how each technique interacts with image structure beyond what numerical metrics alone can reveal.

The LSB watermarked images were visually identical to the originals, as expected. Modifying only the least significant bit introduces negligible perceptual change, reaffirming LSB’s strength in imperceptibility but also hinting at its inherent fragility.

The DCT-based results displayed faint block artifacts in regions containing edges or textual patterns. Although subtle, these distortions highlighted the sensitivity of block transform embedding: altering mid-frequency coefficients unavoidably affects local spatial consistency.

The DWT bit embedding method produced the most noticeable alterations. Images exhibited grain-like distortions and mild contrast shifts due to the global influence of wavelet coefficients. These artifacts suggested that the underlying representation may be too unstable for direct bit replacement, an intuition supported later by the numerical failure rates.

The spread-spectrum methods behaved differently. Spatial SS added a fine noise-like texture reminiscent of film grain: perceptible but not disruptive. DWT-SS, however, offered the cleanest appearance among all non-LSB methods. Embedding the sequence into low-frequency wavelet bands produced only a gentle softening effect, maintaining overall image clarity.  

Together, these observations highlighted a consistent pattern: bit-level embedding excels in invisibility but struggles with stability, while spread spectrum sacrifices minor visual quality to gain robustness.

\begin{table*}[t]
\centering
\small
\caption{Steganography Robustness Performance Across Different Attack Methods.}
\label{tab:attack_comparison}

\setlength{\tabcolsep}{3.5pt}
\renewcommand{\arraystretch}{1.1}

\begin{tabular}{lccccccccccccccccc}
\toprule
\multirow{3}{*}{Method} & \multirow{3}{*}{Total}
& \multicolumn{5}{c}{Gaussian Blur}
& \multicolumn{5}{c}{JPEG Compression (Q=50)}
& \multicolumn{5}{c}{Resize} \\

\cmidrule(lr){3-7} \cmidrule(lr){8-12} \cmidrule(lr){13-17}

& 
& \multicolumn{3}{c}{Success}
& \multicolumn{2}{c}{Failure}
& \multicolumn{3}{c}{Success}
& \multicolumn{2}{c}{Failure}
& \multicolumn{3}{c}{Success}
& \multicolumn{2}{c}{Failure} \\

\cmidrule(lr){3-5} \cmidrule(lr){6-7}
\cmidrule(lr){8-10} \cmidrule(lr){11-12}
\cmidrule(lr){13-15} \cmidrule(lr){16-17}

& 
& Avg & Std & Rate(\%)
& Avg & Std
& Avg & Std & Rate(\%)
& Avg & Std
& Avg & Std & Rate(\%)
& Avg & Std \\

\midrule
LSB        
& 1100 
& 1100.0 & 0.0 & 100.0
& 0.0   & 0.0
& 0.0    & 0.0 & 0.0
& 1100.0 & 0.0
& 1100.0 & 0.0 & 100.0
& 0.0   & 0.0 \\

DCT        
& 1100 
& 779.7  & 2.5 & 70.9
& 320.3 & 2.5
& 0.0    & 0.0 & 0.0
& 1100.0 & 0.0
& 784.1  & 5.3 & 71.3
& 315.9 & 5.3 \\

DWT        
& 1100 
& 795.0  & 3.2 & 72.3
& 305.0 & 3.2
& 0.0    & 0.0 & 0.0
& 1100.0 & 0.0
& 793.8  & 3.6 & 72.2
& 306.2 & 3.6 \\

Spatial SS 
& 1100 
& 1086.8 & 1.7 & 98.8
& 13.2  & 1.7
& 1065.0 & 1.8 & 96.8
& 35.0  & 1.8
& 1085.8 & 3.5 & 98.7
& 14.2  & 3.5 \\

DWT-SS     
& 1100 
& 1080.9 & 2.0 & 98.3
& 19.1  & 2.0
& 851.3  & 3.9 & 77.4
& 248.7 & 3.9
& 1081.7 & 3.4 & 98.3
& 18.3  & 3.4 \\

TrustMark  
& 1100 
& 368.4  & 2.7 & 33.5
& 731.6 & 2.7
& 608.4  & 3.2 & 55.3
& 491.6 & 2.4
& 902.3  & 2.9 & 82.0
& 197.7 & 2.9 \\

\bottomrule
\end{tabular}
\end{table*}

\begin{figure*}[ht]
\centering
\small
\setlength{\tabcolsep}{2.5pt}
\renewcommand{\arraystretch}{1.12}

\textbf{Image caption: “Shine a laser pointer at it”}

\begin{tabular}{|m{2.5cm}|c|c|c|c|}
\hline
\textbf{} & \textbf{Original(encode)} & \textbf{Gaussian Blur} & \textbf{JPEG Compression (Q=50)} & \textbf{Resize} \\
\hline

\textbf{Original image} &
\includegraphics[height=1.8cm]{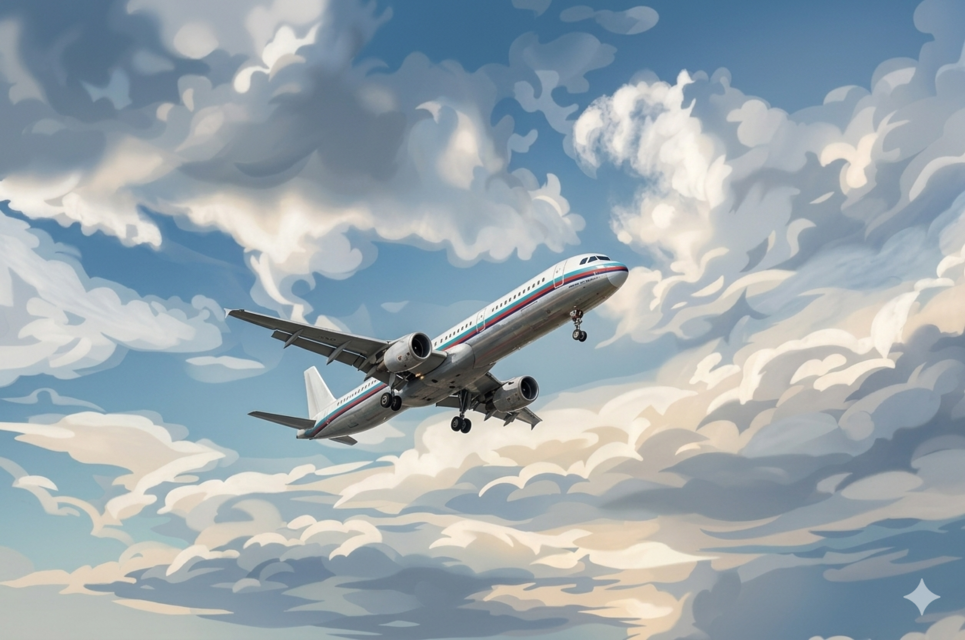} &
\includegraphics[height=1.8cm]{figure/1095.png} &
\includegraphics[height=1.8cm]{figure/1095.png} &
\includegraphics[height=1.8cm]{figure/1095.png} \\
\hline

\textbf{TrustMark} &
\includegraphics[height=1.8cm]{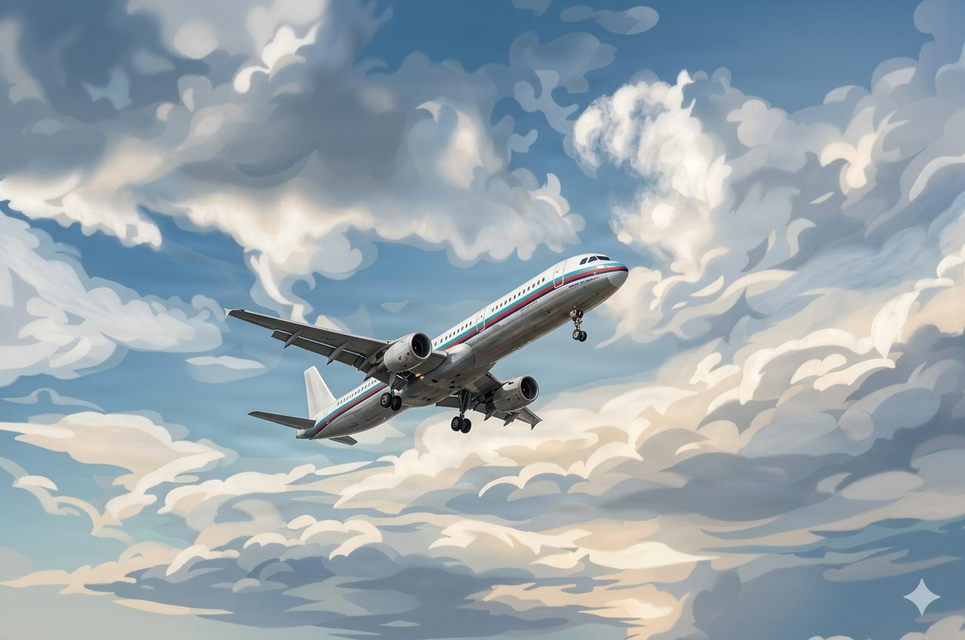} &
\includegraphics[height=1.8cm]{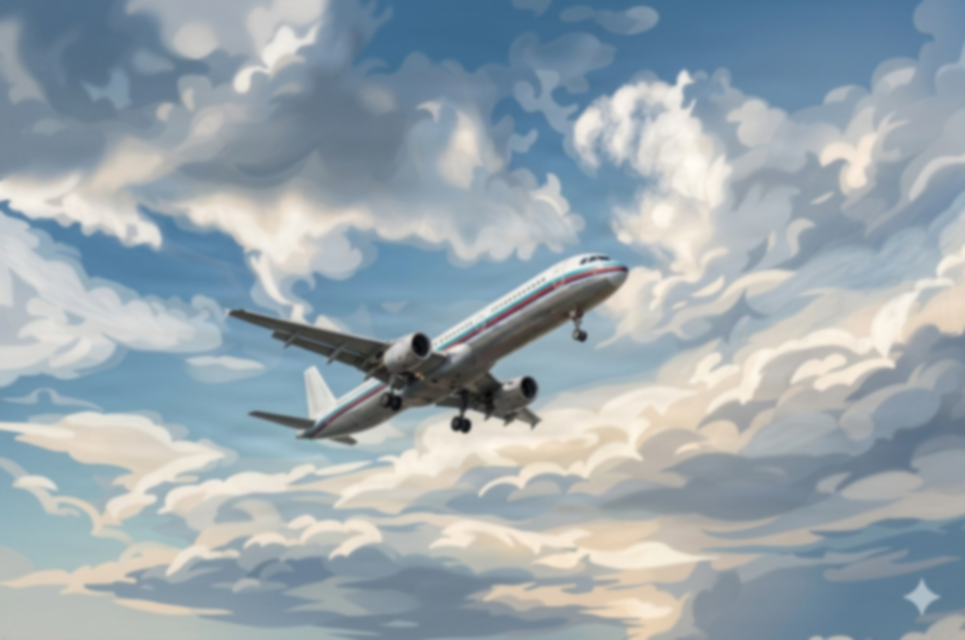} &
\includegraphics[height=1.8cm]{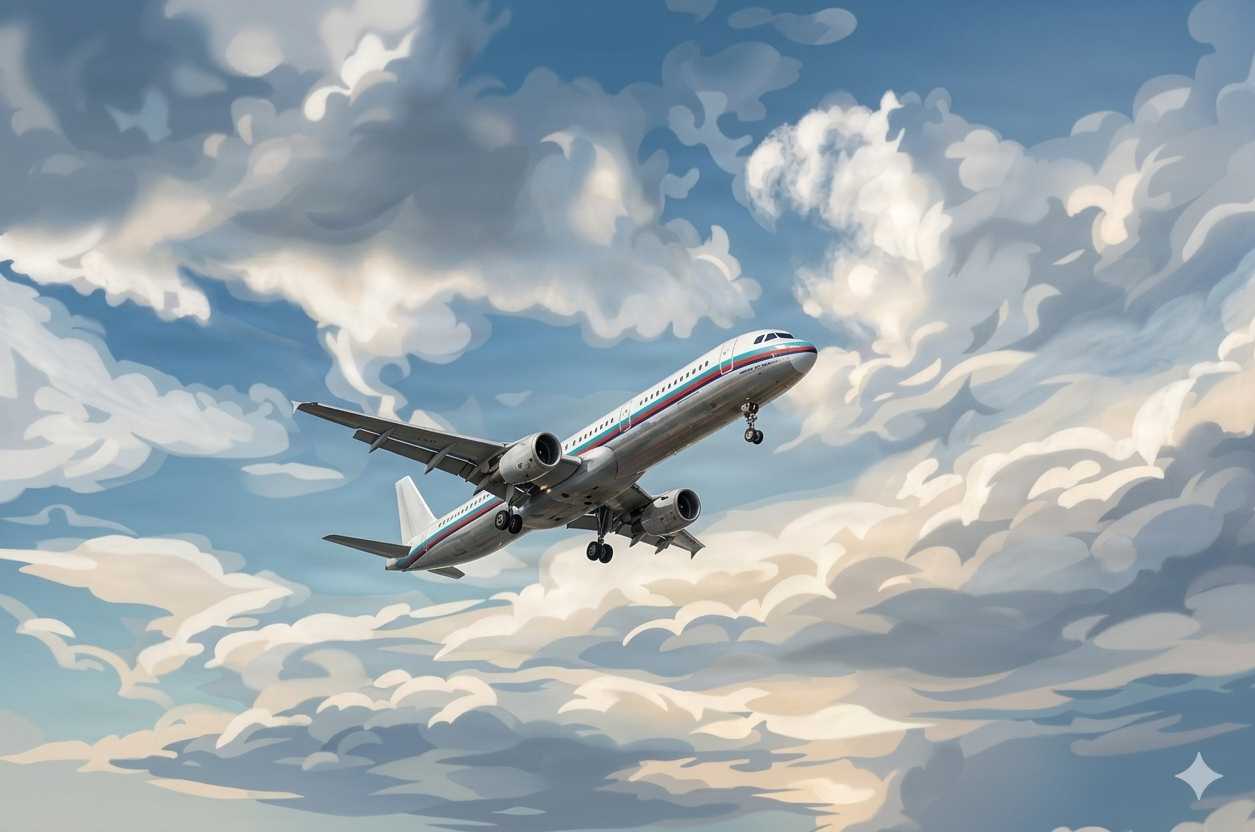} &
\includegraphics[height=1.8cm]{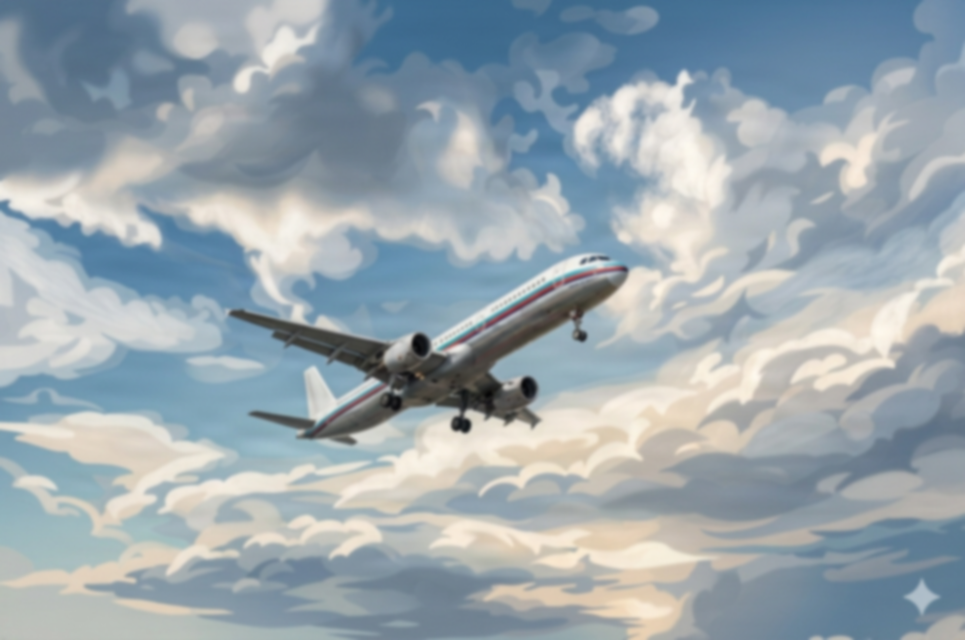} \\
\hline

\textbf{LSB} &
\includegraphics[height=1.8cm]{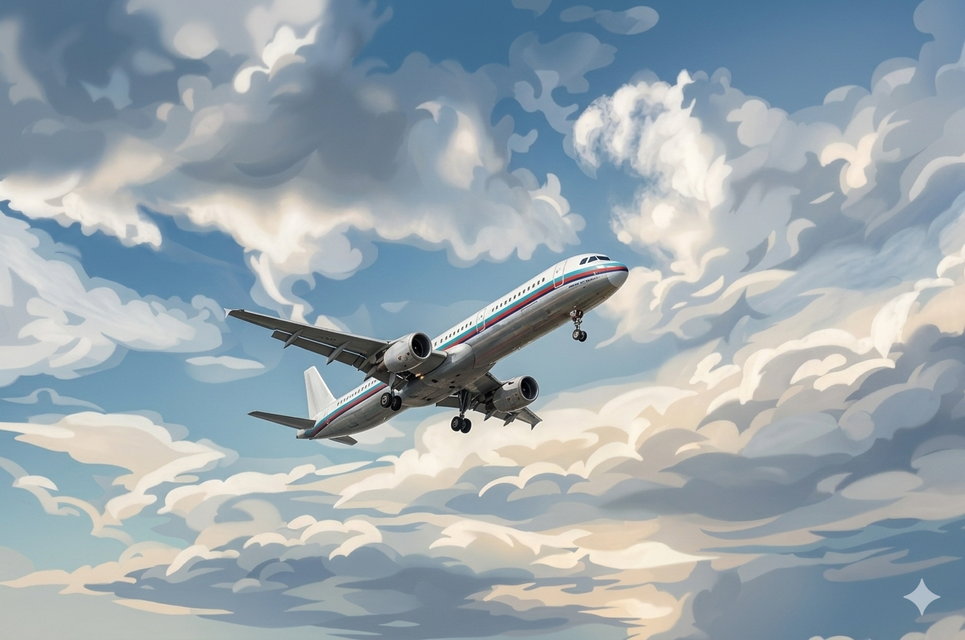} &
\includegraphics[height=1.8cm]{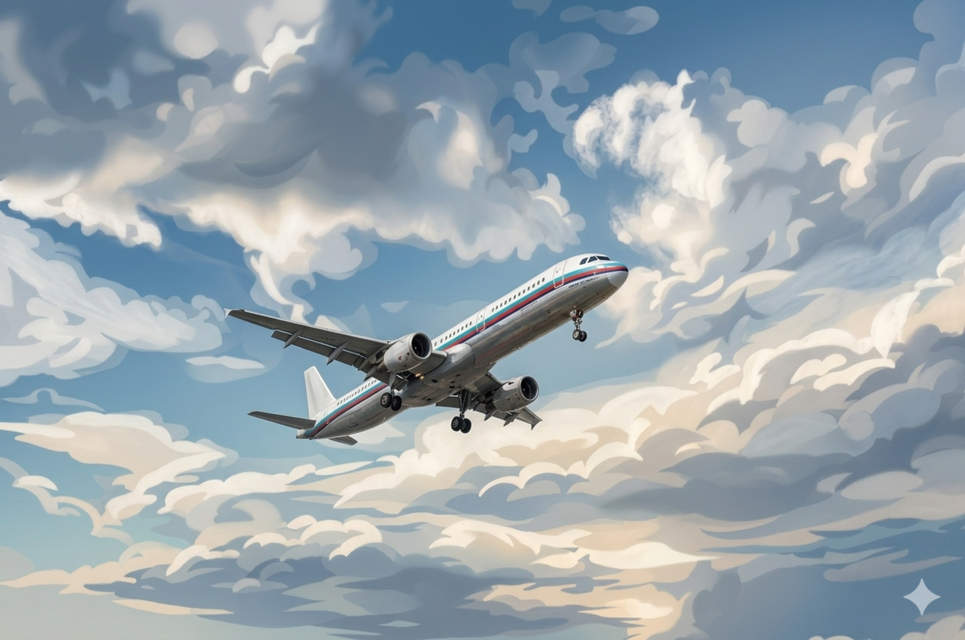} &
\includegraphics[height=1.8cm]{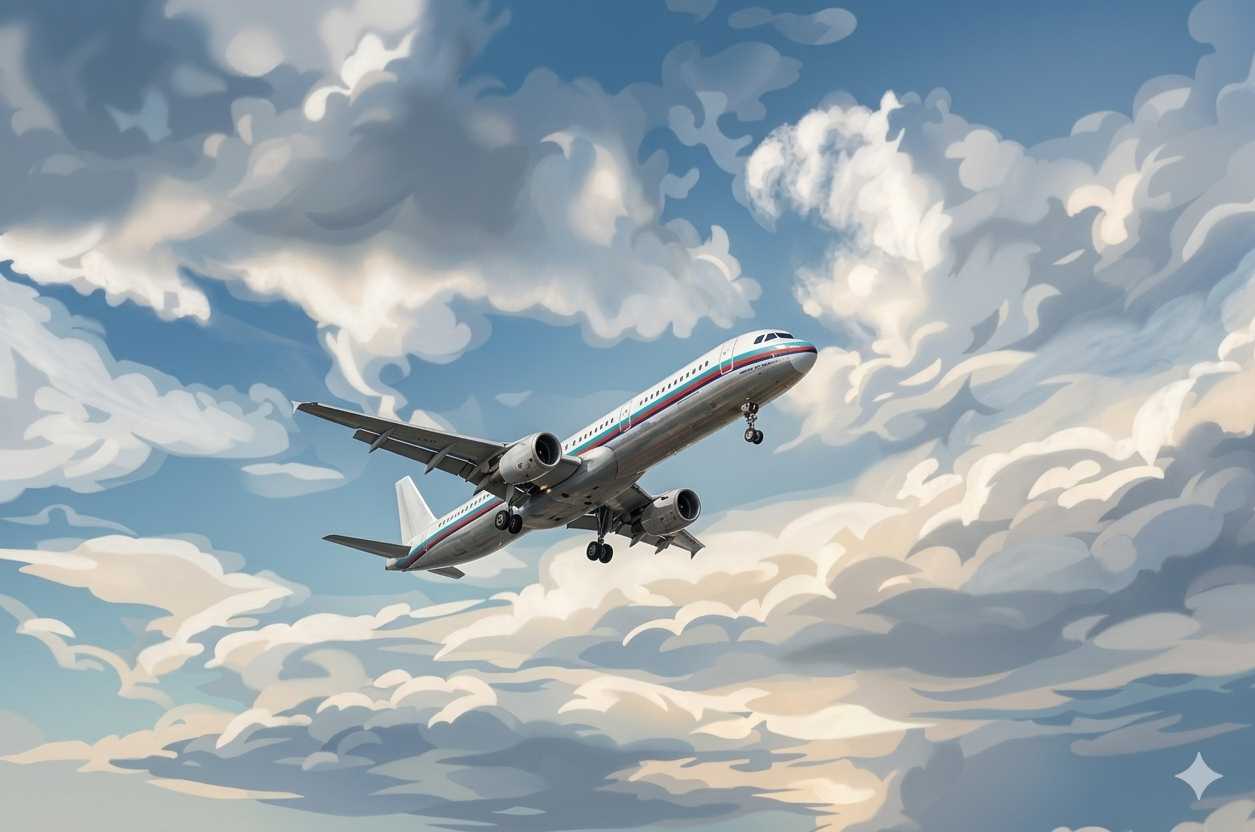} &
\includegraphics[height=1.8cm]{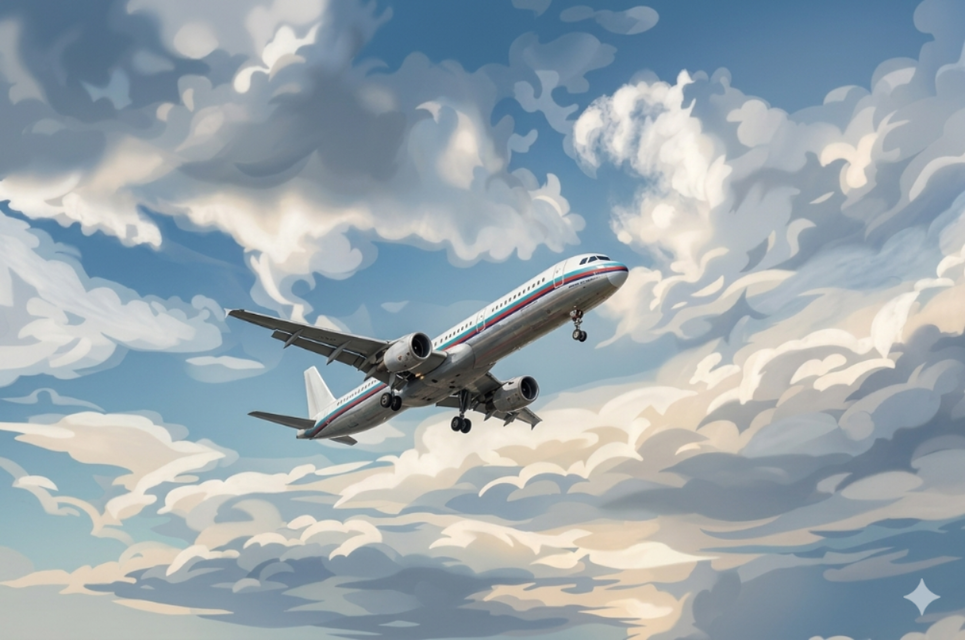} \\
\hline

\textbf{DCT} &
\includegraphics[height=1.8cm]{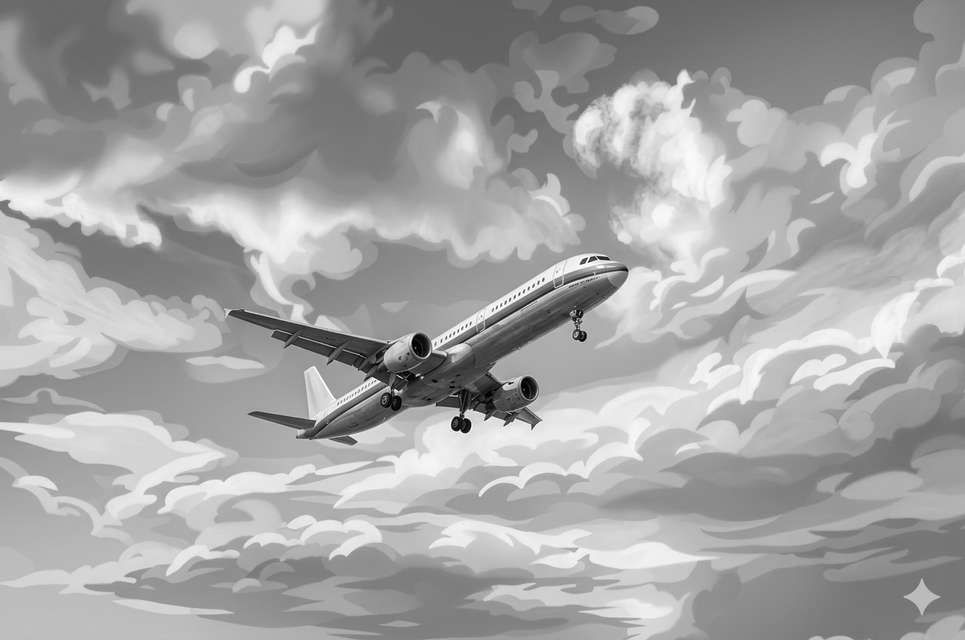} &
\includegraphics[height=1.8cm]{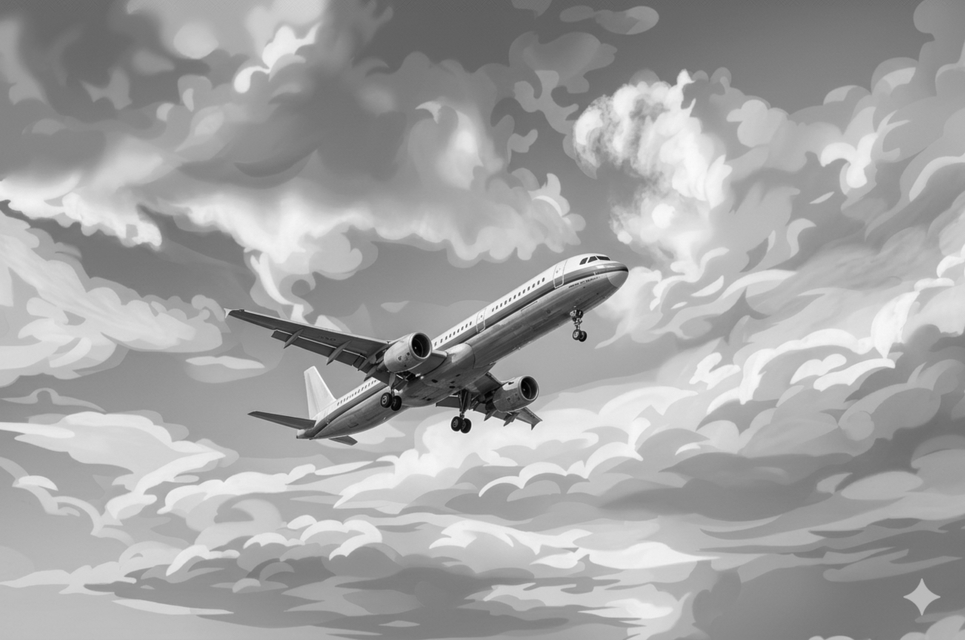} &
\includegraphics[height=1.8cm]{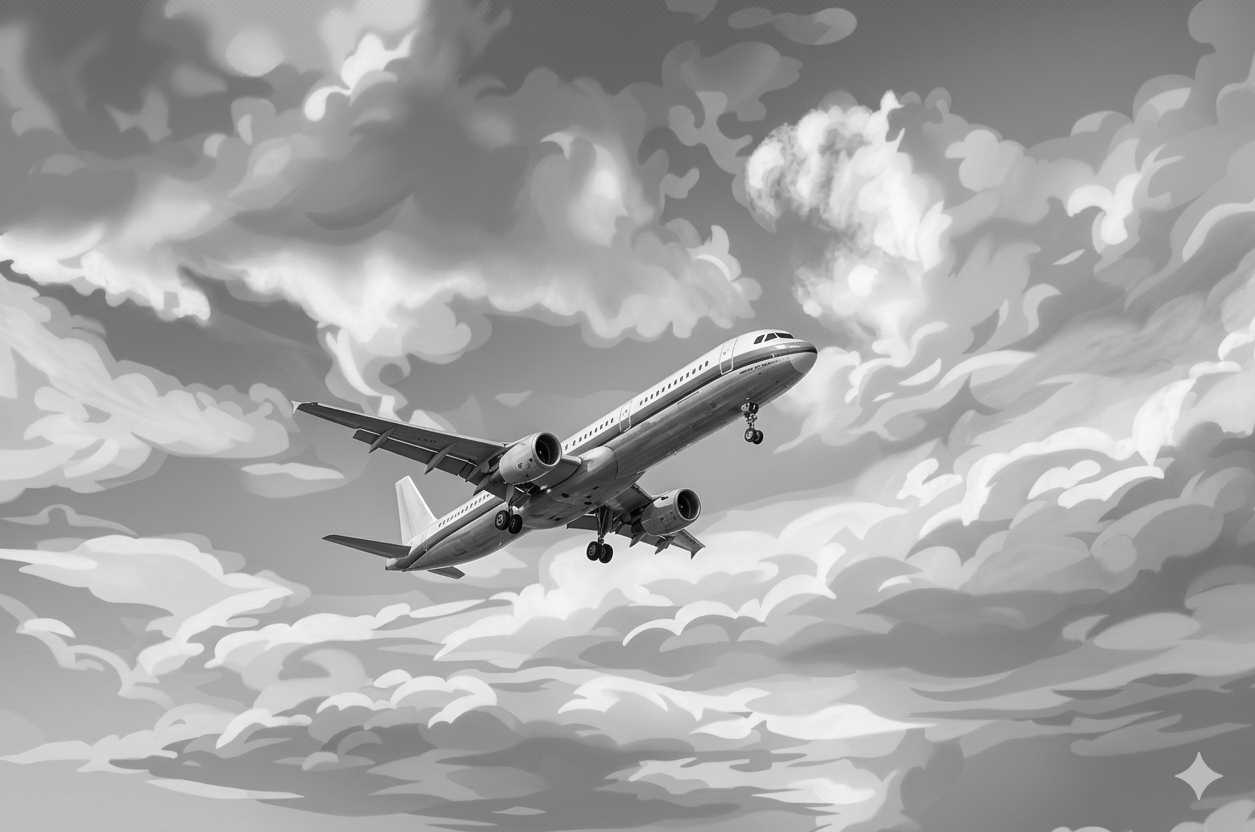} &
\includegraphics[height=1.8cm]{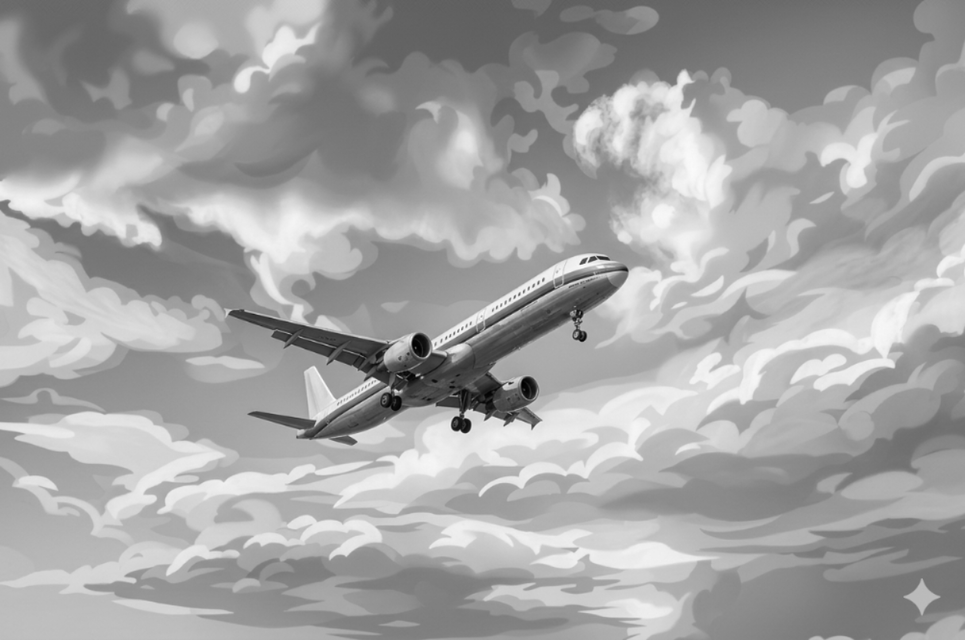} \\
\hline

\textbf{DWT} &
\includegraphics[height=1.8cm]{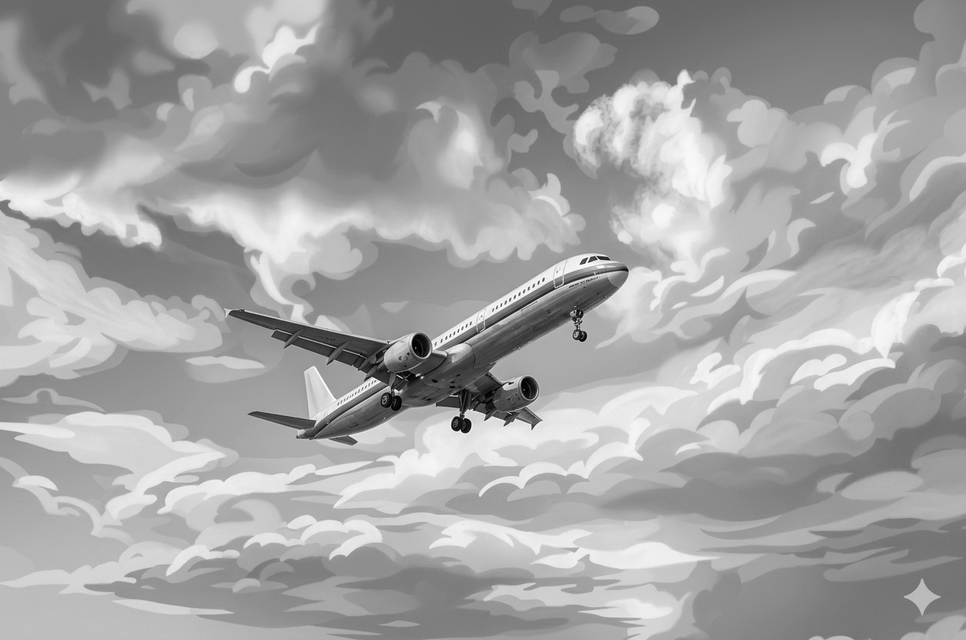} &
\includegraphics[height=1.8cm]{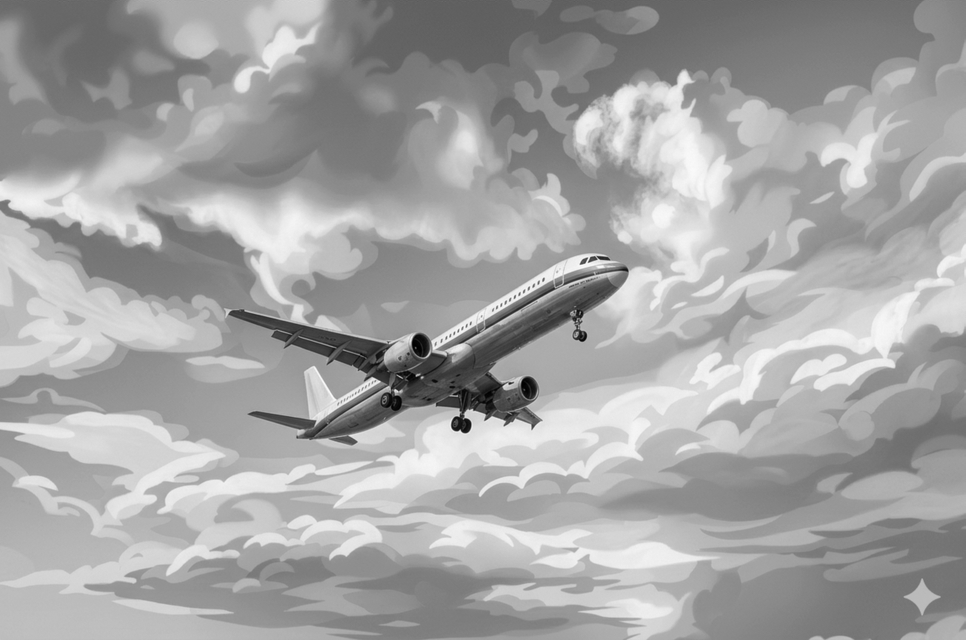} &
\includegraphics[height=1.8cm]{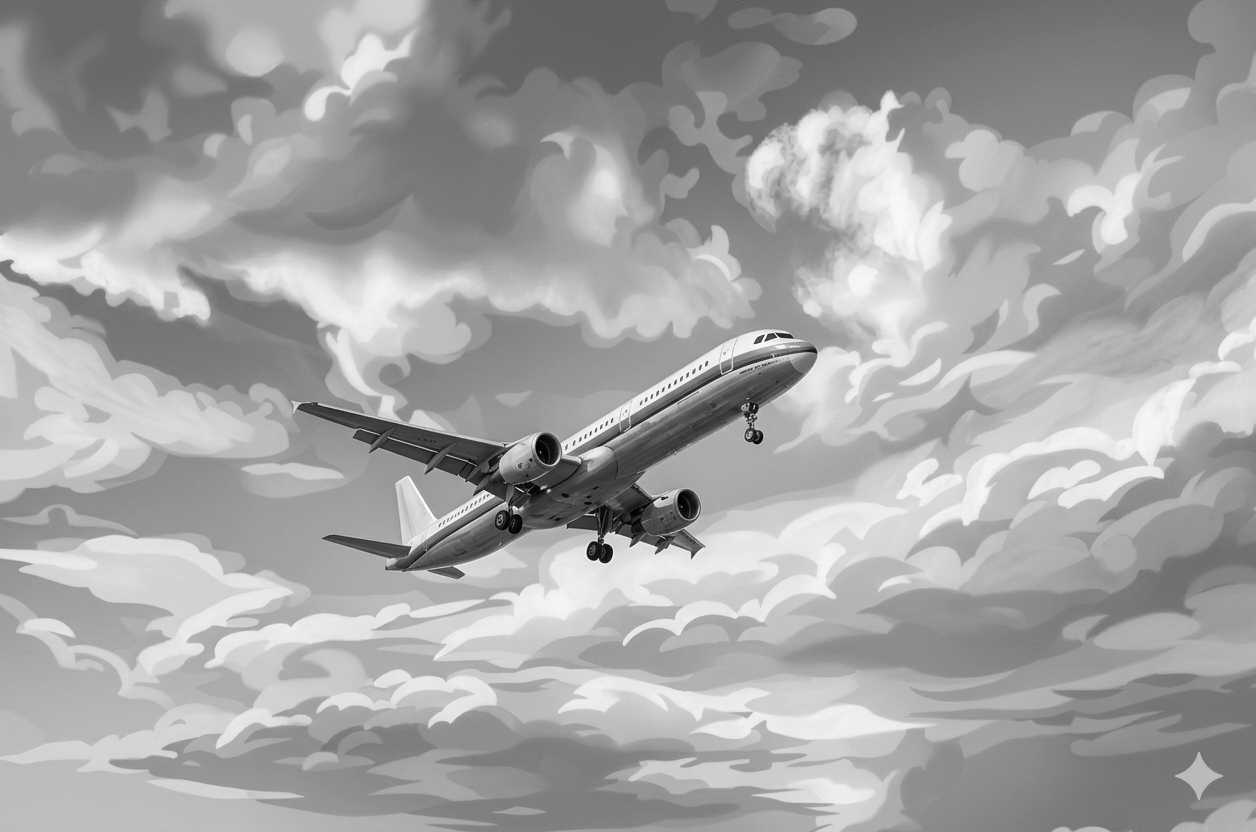} &
\includegraphics[height=1.8cm]{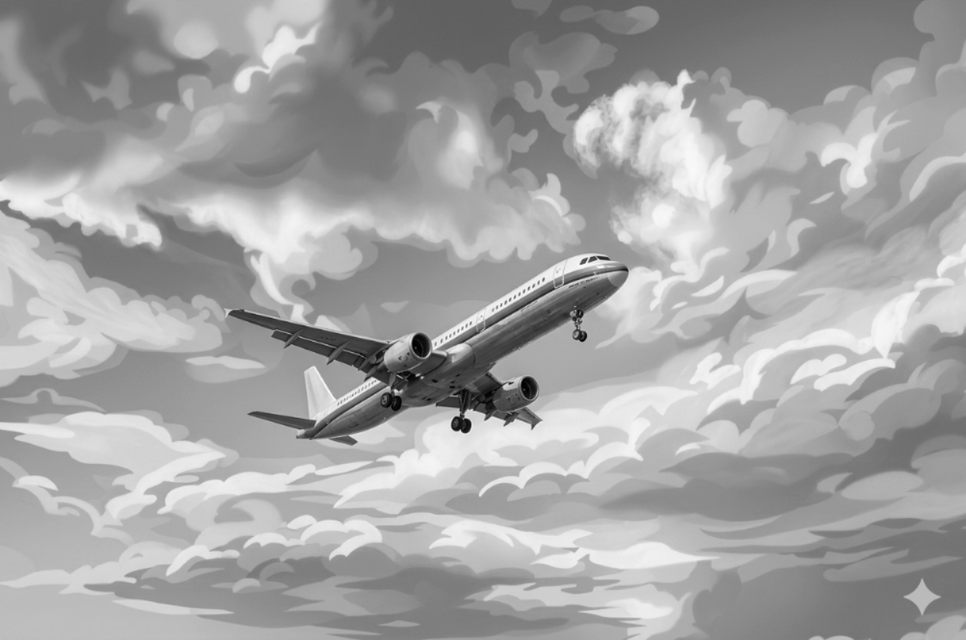} \\
\hline

\textbf{Spatial SS} &
\includegraphics[height=1.8cm]{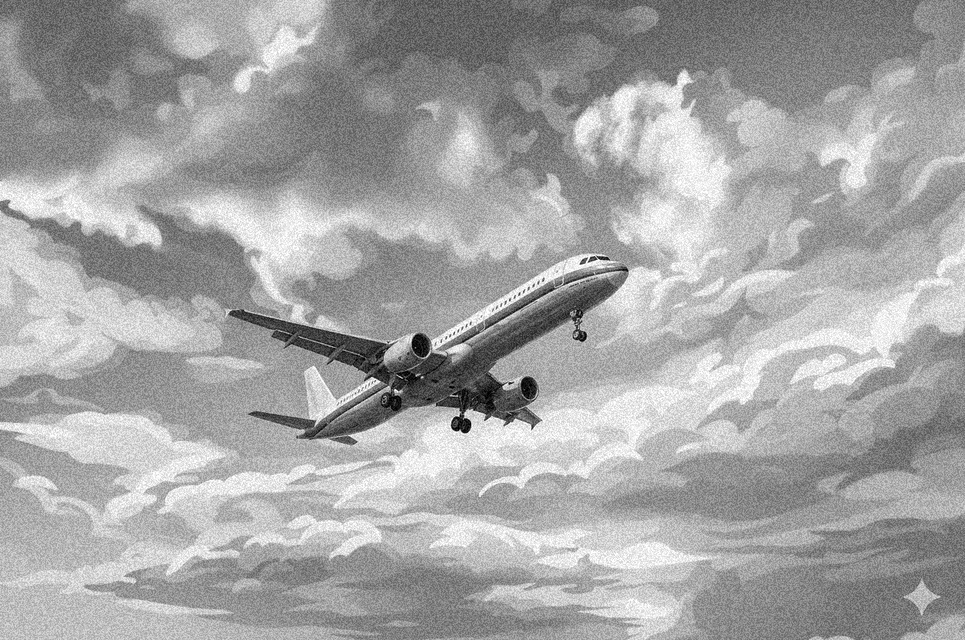} &
\includegraphics[height=1.8cm]{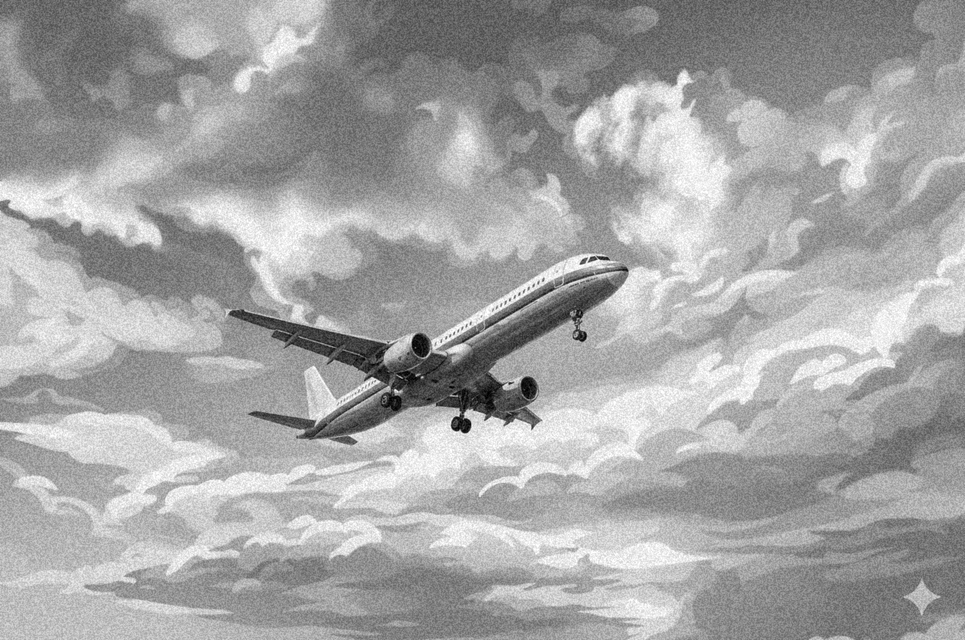} &
\includegraphics[height=1.8cm]{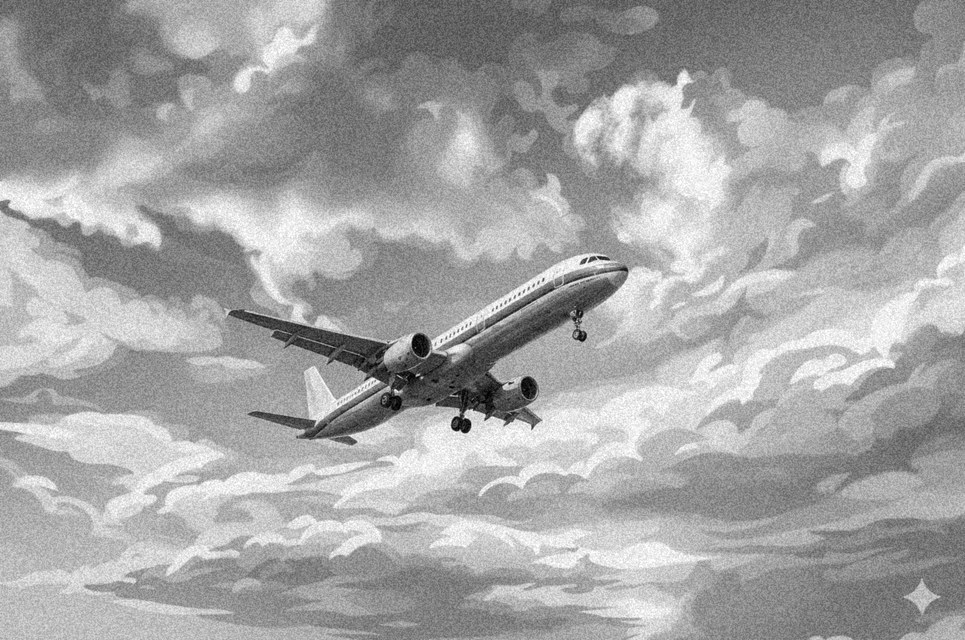} &
\includegraphics[height=1.8cm]{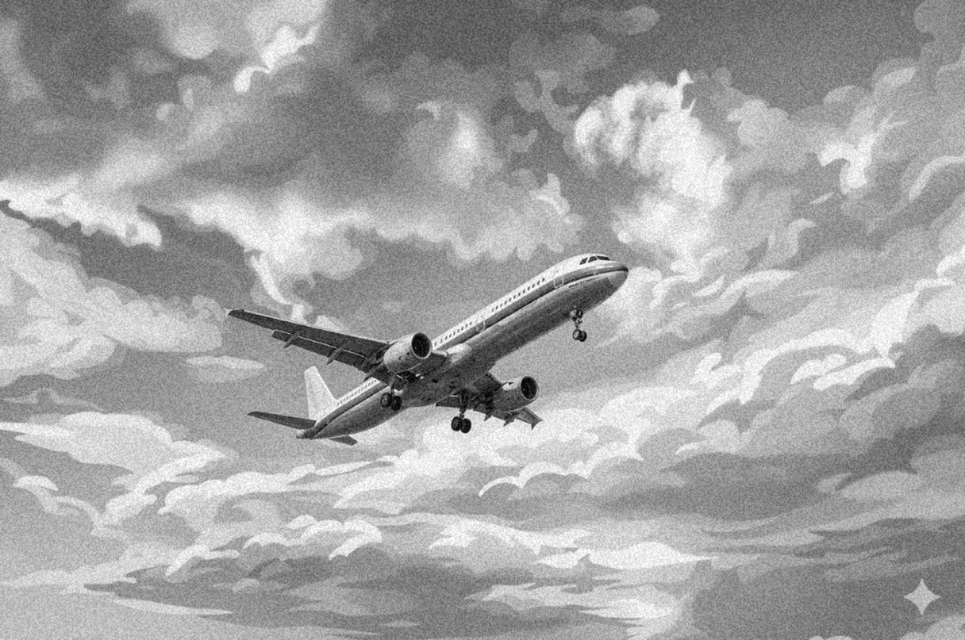} \\
\hline

\textbf{DWT-SS} &
\includegraphics[height=1.8cm]{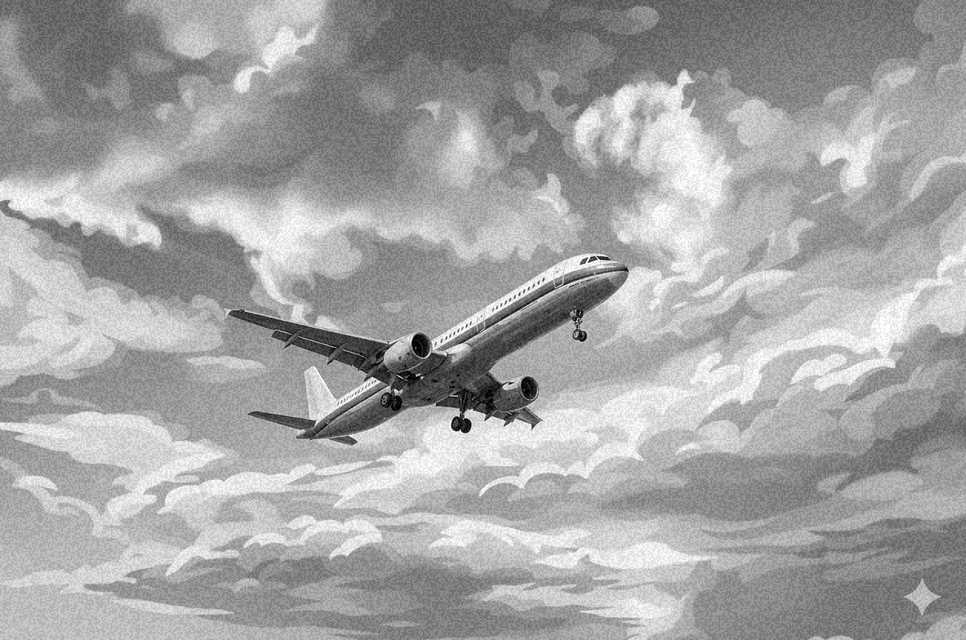} &
\includegraphics[height=1.8cm]{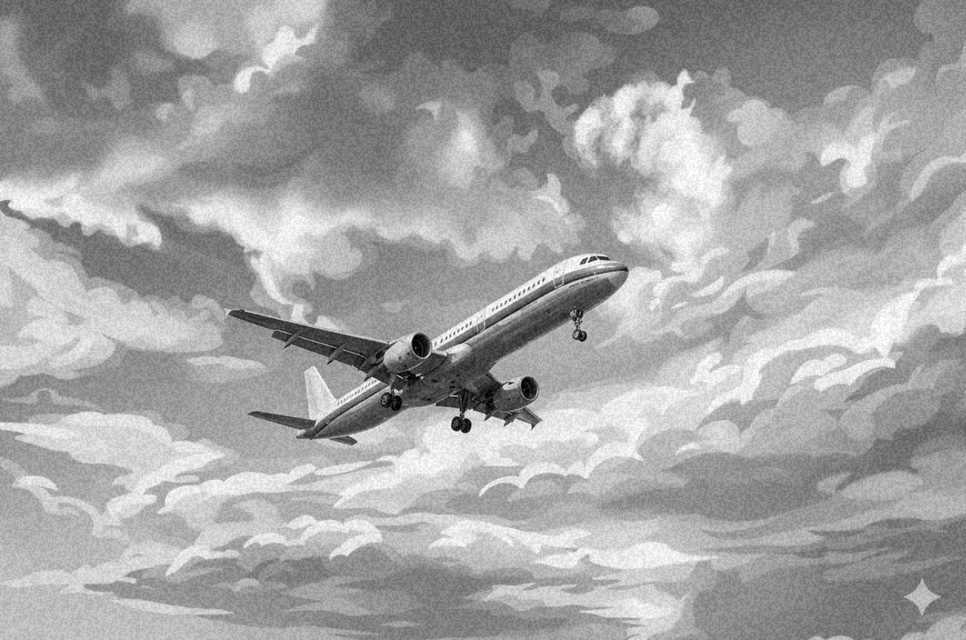} &
\includegraphics[height=1.8cm]{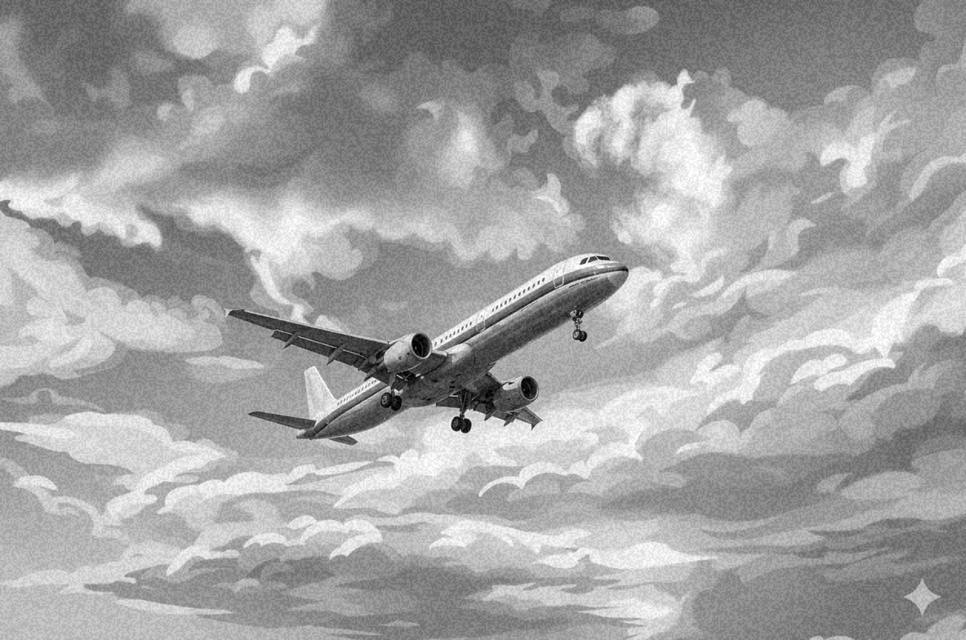} &
\includegraphics[height=1.8cm]{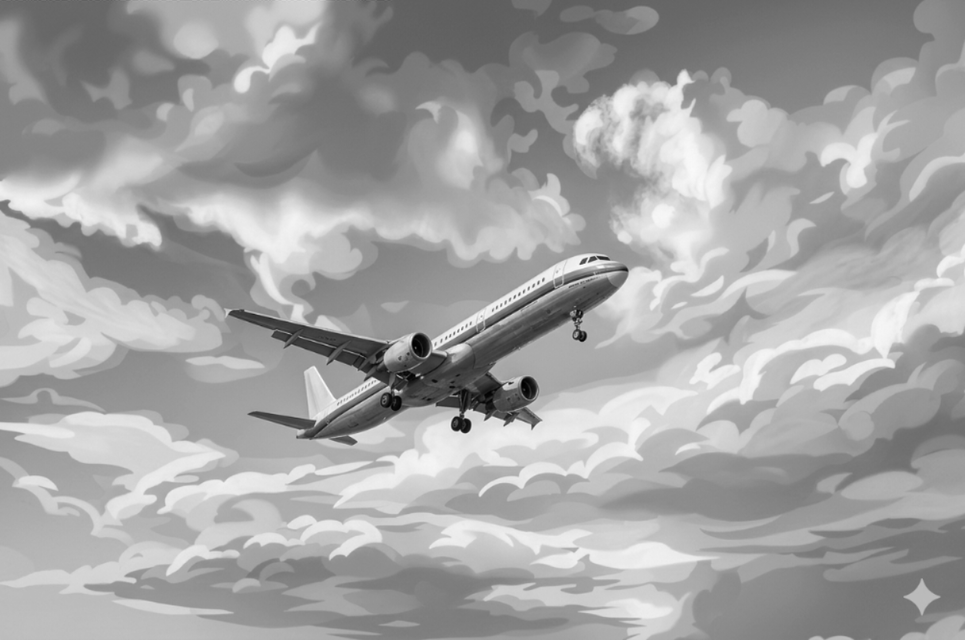} \\
\hline

\end{tabular}

\caption{Visual comparison of watermarking methods on a natural color image under different attack conditions.}

\label{fig:plane}
\end{figure*}

\subsection{Effectiveness and Robustness of Steganography}

This subsection provides a unified comparative evaluation under identical experimental conditions, which we use as a robustness benchmark across different watermarking approaches.

The quantitative evaluation confirmed the qualitative trends,
as summarized in Table~\ref{tab:attack_comparison}. All reported values are averaged over 10 independent runs, with standard deviation (Std) indicating variability across repeated trials. Under no attack conditions, LSB performed perfectly, and every RSA signature was recovered without error. This is consistent with the expectation that bit substitution works reliably as long as the host image remains unchanged.

In contrast, the DCT method produced 320 failures across the dataset. Even minor variations caused by coefficient rounding or slight shifts in block boundaries were sufficient to break the exact bit reconstruction required for RSA verification. The DWT bit embedding method showed moderate fragility as well, resulting in 305 failures. This reflects the sensitivity of wavelet domain coefficients, especially under floating-point operations and multi-level decomposition, where small numerical changes can propagate and lead to decoding errors.
  
In comparison, the spread spectrum schemes delivered much more stable results. Spatial SS exhibited only 13 failures, while DWT SS showed 19 failures, making both methods orders of magnitude more reliable than bit-level approaches. Their correlation-based detection tolerates natural image variation, which explains their stronger robustness under realistic conditions.

From a statistical perspective, the relatively low standard deviation values, typically between 1.7 and 3.2, indicate that the observed performance is consistent across repeated runs. LSB shows zero variance, confirming perfectly stable decoding behavior. DCT and DWT exhibit slightly higher variability, which reflects their sensitivity to embedding perturbations. In contrast, the spread spectrum methods maintain both high success rates and low variance, further supporting their robustness and reproducibility.

The blur attack stage further amplified these differences. All bit-level methods, including LSB, DCT, and DWT, collapsed entirely after Gaussian blur, producing zero successful recoveries. Even minimal smoothing perturbs pixel values enough to destroy exact RSA bit reconstruction. In contrast, the spread spectrum methods remained robust. Spatial SS and DWT SS each recorded only 19 failures under blur, which is only slightly worse than their no attack performance. DWT SS in particular maintained low bit error rates in the range of 0.031 to 0.219, indicating that low-frequency wavelet embedding is naturally resilient to smoothing operations.

In addition to the steganographic techniques examined above, TrustMark was included in the evaluation as a representative attribution-oriented watermarking method, as shown in Figures~\ref{fig:plane}. TrustMark differs from the previous approaches in its detection philosophy. Rather than relying on exact bit recovery or correlation thresholds, it focuses on whether an embedded payload can be detected and whether the decoded message matches the original signature. This design places TrustMark closer to semantic level verification than traditional signal level watermarking.

For consistency, TrustMark was evaluated using the same image set and the same blur perturbation applied in earlier experiments. All images were first processed by the official TrustMark encoder and then decoded under both pristine and blurred conditions. Three indicators were considered. Decode success reflects whether the decoder produces a valid output. Watermark presence indicates whether a watermark signal is detected. Payload accuracy measures whether the recovered message matches the embedded signature.

On clean encoded images, TrustMark showed stable behavior. A watermark was detected in 942 out of 1097 images, corresponding to a presence rate of 85.9 percent. In every detected case, the decoded payload exactly matched the original signature. No decoding failures were observed, suggesting that the extraction process is reliable when images remain unaltered.

After applying the Gaussian blur with radius 0.5, TrustMark remained detectable in 373 images, which is 34.0 percent of the dataset. Among these detections, 366 images successfully recovered the correct payload, resulting in an overall payload accuracy of 33.4 percent. Notably, the decoder still produced valid outputs for all images. This indicates that the performance degradation is caused by weakened watermark signals rather than decoding instability.

Collectively, these findings indicate that Gaussian blur has a significant effect on the detectability of TrustMark. However, the degradation occurs gradually rather than as a sudden failure. Even though blur weakens the embedded watermark, TrustMark does not entirely break down under this common image perturbation, and a substantial fraction of watermarks remains recoverable.

Beyond blur-based perturbations, we further examine how the watermarking schemes behave under compression and geometric transformations, which are common in real-world image pipelines.

As summarized in Table~\ref{tab:attack_comparison}, JPEG compression at Q=50 exposes a clear robustness gap between bit-level and correlation-based embedding. All reported values are averaged over 10 independent runs, with standard deviation (Std) reflecting variability across repeated trials.
Bit-level methods (LSB/DCT/DWT) rely on exact reconstruction of embedded bits; however, JPEG introduces quantization and rounding that alter pixel values and frequency coefficients, thereby breaking deterministic bit recovery and leading to verification collapse.
In contrast, spread-spectrum schemes remain substantially more stable because detection is performed via correlation over many samples, which tolerates moderate quantization noise.
Within the spread-spectrum family, Spatial SS achieves higher verification success than DWT SS at Q=50, suggesting that the chosen spatial domain correlation structure and its associated synchronization are better matched to JPEG-induced distortions in our setting.

From a statistical perspective, the standard deviation values under JPEG compression remain relatively low, for example, 1.8 for Spatial SS and 3.9 for DWT SS, indicating stable performance across repeated runs despite compression-induced perturbations. The slightly higher variance observed in DWT SS reflects its increased sensitivity to quantization noise compared to Spatial SS.

Table~\ref{tab:attack_comparison} highlights a complementary failure mode under geometric resampling, specifically 80 percent resize followed by center crop. All results are likewise averaged over 10 runs.
Resizing introduces interpolation and spatial misalignment, which degrades transform-domain approaches, such as DCT with 780 out of 1100 and DWT with 795 out of 1100, and reduces the effective detectability of structured embeddings.
Nevertheless, spread-spectrum approaches remain highly robust, with Spatial SS achieving 1087 out of 1100 and DWT SS achieving 1081 out of 1100. This is consistent with the intuition that globally distributed correlation energy is less sensitive to local resampling artifacts than localized coefficient perturbations.

The standard deviation values under resizing, generally within 2.9 to 5.3, indicate slightly higher variability compared to JPEG compression, which is expected given the additional spatial misalignment introduced by interpolation. However, the spread-spectrum methods still maintain both high success rates and moderate variance, further demonstrating their robustness and reproducibility.

TrustMark exhibits noticeable degradation under resizing, with 903 out of 1097, but does not collapse entirely, indicating partial robustness under geometric distortions. We leave a quantitative TrustMark evaluation under JPEG compression to future work for a fully consistent cross-attack comparison.

\subsubsection{Demonstration of Steganographic Samples}

Bit-level methods preserve visual quality, while spread-spectrum methods introduce mild noise-like artifacts, as illustrated in Figure~\ref{fig:plane}. The differences can be clearly observed in the visual comparison without any attacks. LSB is the most loyal to the visual appearance: LSB is the only technique that preserves the original color palette almost exactly, which means that the watermarked image is almost similar to the original. DCT exhibits light block artifacts on edges, indicating the sensitivity of mid-frequency coefficients of DCT to quantization. DWT bit embedding, on the other hand, introduces the most visible distortions. Because its modifications propagate across wavelet levels, the resulting image appears slightly grainy and carries a noticeable shift in brightness.

The two spread-spectrum methods maintain much better perceptual quality. Spatial SS adds a soft, film-like noise pattern, but the overall structure and content remain clear. DWT-SS achieves the cleanest output among the non-LSB approaches. By placing the watermark in low-frequency wavelet bands, it produces only a gentle smoothing effect, allowing the image to stay visually close to the original while still embedding a detectable signal.

Under Gaussian blur, the contrast between methods becomes even sharper, as shown in Figure~\ref{fig:plane}.
 LSB, DCT, and DWT all lose their watermark information completely: the blur suppresses the fine-grained pixel variations they rely on, leaving an image that looks visually acceptable but carries no recoverable watermark. Spatial SS fares better. Even though blur weakens its high-frequency components, its low-frequency correlation structure survives, which matches the relatively low BER values observed in the experiments.
DWT-SS proves to be the most robust overall. After blurring, the image still retains its structure, texture, and watermark response, with only mild softening. This stability highlights the strength of embedding information in low-frequency wavelet components, which tend to withstand smoothing and filtering operations much more effectively.
Taken together, the visual analysis supports a clear conclusion:
Bit-level embedding excels at imperceptibility, especially LSB, which uniquely preserves color but collapses under distortions, while spread spectrum approaches sacrifice a bit of visual purity to gain significantly higher robustness. Among them, DWT-SS provides the most well-balanced performance.

\subsection{Harmful Detection Training and Evaluation}

\begin{figure}
  \centering
  \fbox{
    \includegraphics[width=0.9\linewidth]{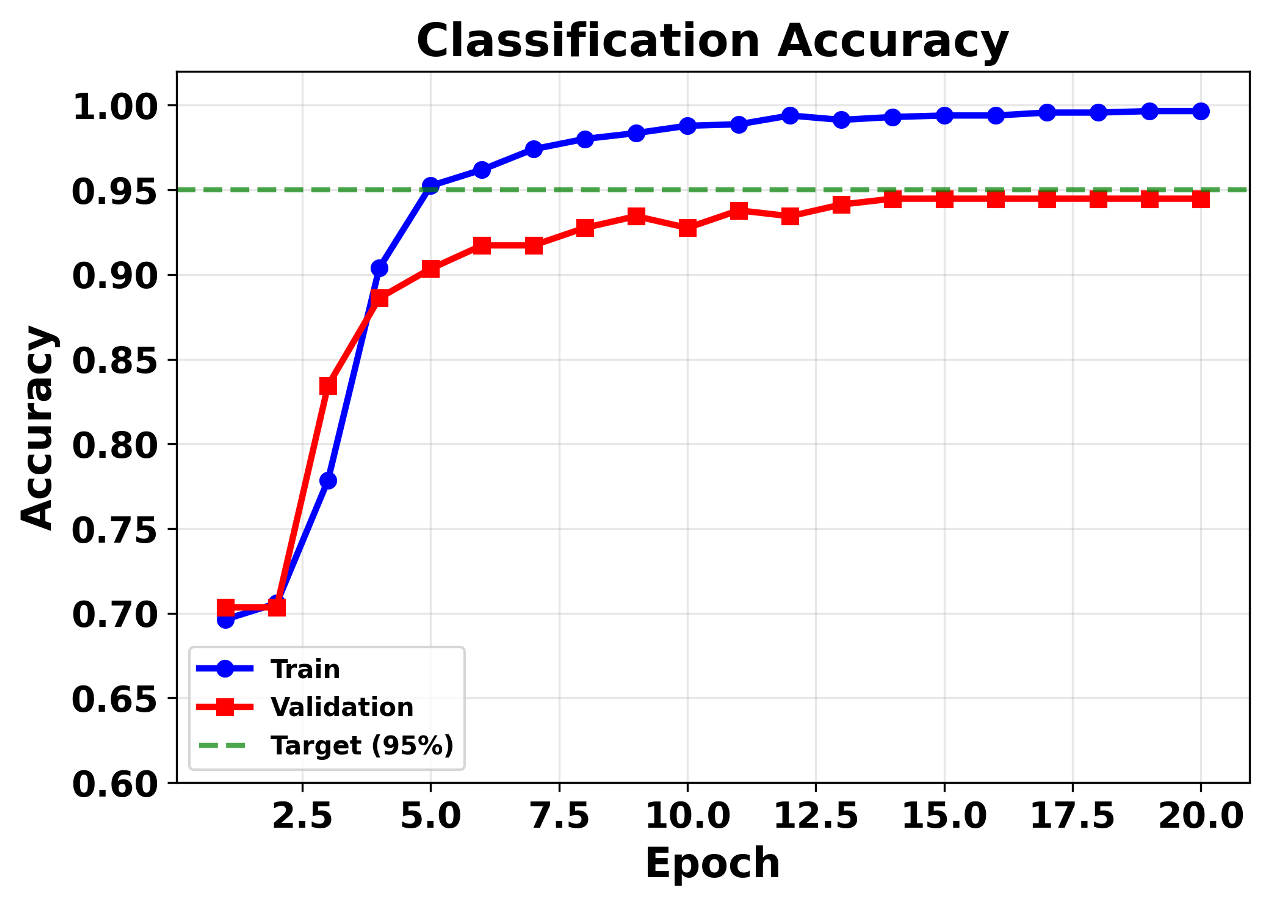}
  }
  \caption{Training Accuracy and Validation Performance.}
  \label{fig:exp:training_acc}
\end{figure}


To evaluate the effectiveness of our multimodal fusion classifier, we train our model on paired image–text samples and analyze its learning behavior across epochs. 

Figure~\ref{fig:exp:training_acc} shows the training and validation accuracy. Loss curves decrease smoothly, and validation accuracy tracks closely with the training curve, approaching 95\%. The small and stable gap between the two sets of metrics suggests minimal overfitting and consistent agreement between training and validation performance. Overall, these results demonstrate that our multimodal fusion classifier trains stably, exhibits low overfitting, and achieves strong validation performance.

\section{Conclusion}
A current limitation lies in the sensitivity of transform-domain bit embedding to numerical precision. RSA verification requires exact bit-level reconstruction of the embedded signature, yet inverse transforms inevitably introduce small rounding and floating-point variations. Although we mitigated this through standardized quantization steps, consistent rounding rules, and stricter block alignment, the fundamental tension between lossy signal processing and bit-exact cryptographic verification remains. Under severe distortions beyond those evaluated here, verification failures may still occur. Future work will explore error-tolerant signature schemes and soft-decision decoding to relax the bit-exactness requirement.

This work introduced STAMP-V, a forensic framework that embeds cryptographically signed identifiers into images and verifies their provenance through multimodal analysis. Experiments across five watermarking schemes showed that spread-spectrum methods, particularly DWT-SS, provide strong robustness, while our CLIP-based fusion classifier achieves near-perfect detection performance. Together, these components form a practical and reliable foundation for provenance verification and harmful-content detection in AI-generated imagery.


{
    \small
    \bibliographystyle{IEEEtran}
    \bibliography{references}
}

\section{Biography Section}

\begin{IEEEbiography}[{\includegraphics[width=0.9in,height=1.1in,clip,keepaspectratio]{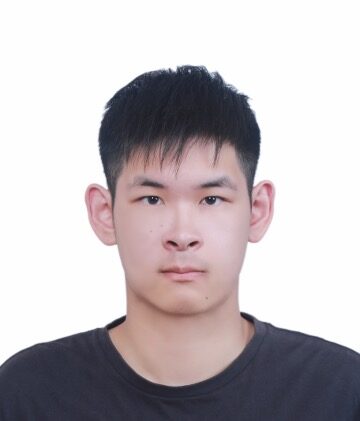}}]{Xinlei Guan}
Xinlei Guan is currently pursuing the B.S. degree in computer science at Kean University, Union, NJ, USA. His research interests include AI-generated content security, multimodal learning, and steganography.
\end{IEEEbiography}

\begin{IEEEbiography}[{\includegraphics[width=0.9in,height=1.1in,clip,keepaspectratio]{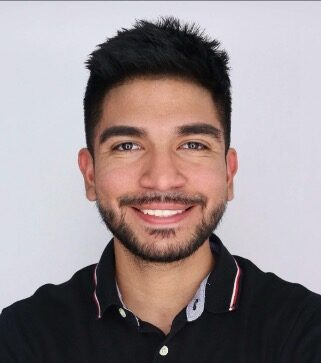}}]{David Arosemena}
David Arosemena is currently pursuing the B.S. degree in computer science with a concentration in cybersecurity at Kean University, Union, NJ, USA. His research interests include image watermarking, steganography, multimedia provenance, and AI-based content verification.
\end{IEEEbiography}

\begin{IEEEbiography}[{\includegraphics[width=0.9in,height=1.1in,clip,keepaspectratio]{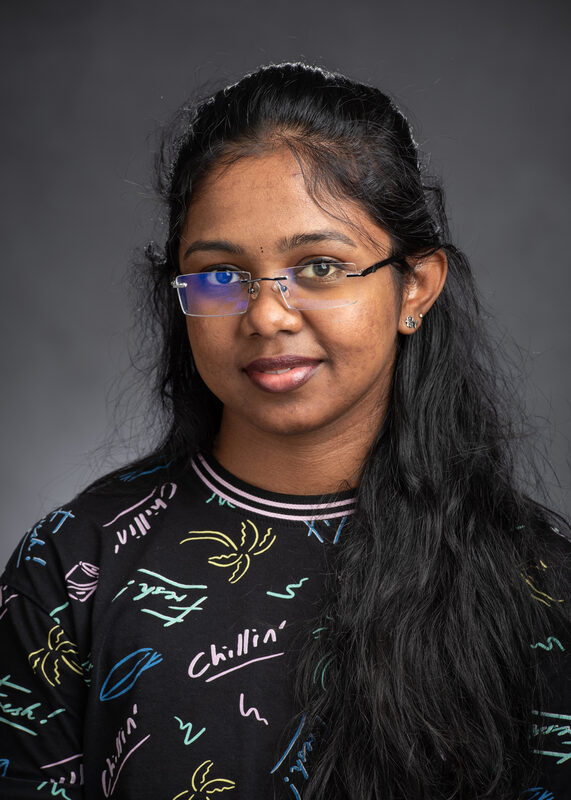}}]{Tejaswi Dhandu}
Tejaswi Dhandu is a Ph.D. student in Electrical and Computer Engineering at North Dakota State University, Fargo, ND, USA. She received the M.S. degree in Electrical Engineering from the University of Missouri--Kansas City, USA. Her research focuses on thermal-aware architecture, VLSI design, hardware--AI systems, and trustworthy machine learning.
\end{IEEEbiography}

\begin{IEEEbiography}[{\includegraphics[width=0.9in,height=1.1in,clip,keepaspectratio]{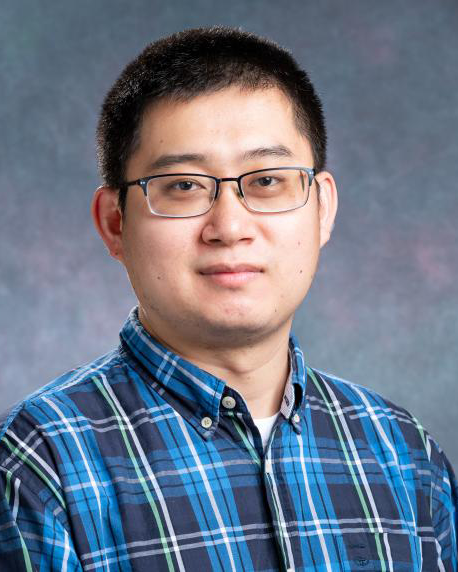}}]{Kuan Huang}
Kuan Huang (Member, IEEE) received the B.Eng. degree in electrical engineering and automation from Harbin Institute of Technology, Harbin, China, in 2016 and the Ph.D. degree in computer science from Utah State University, Logan, UT, USA, in 2021. Since 2022, he has been an Assistant Professor with the Department of Computer Science and Technology, Kean University, Union, NJ, USA. His research focuses on computer vision, deep learning, and medical image analysis.
\end{IEEEbiography}

\begin{IEEEbiography}[{\includegraphics[width=0.9in,height=1.1in,clip,keepaspectratio]{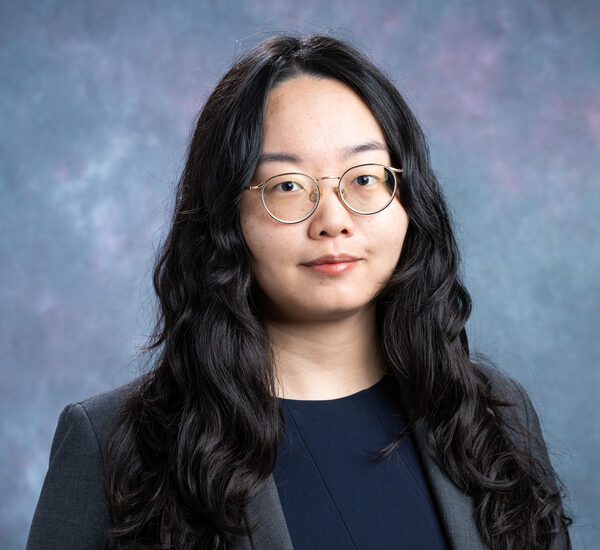}}]{Meng Xu}
MENG XU (Member, IEEE) received the B.S. degree in Management Information Systems from Tianjin University of Technology, Tianjin, China, in 2017, and the Ph.D. degree in Computer Science from Utah State University, Logan, UT, USA, in 2023.
She is currently an Assistant Professor with the Department of Computer Science and Technology at Kean University, Union, NJ, USA. Her research focuses on computer vision, deep learning, and medical image analysis.

Dr. Xu’s research has been supported by multiple grants and programs, including the NSF CISE MSI program, the AIM-AHEAD Research Fellowship, and the CAHSI–Google Institutional Research Program (IRP).
\end{IEEEbiography}

\begin{IEEEbiography}[{\includegraphics[width=0.9in,height=1.1in,clip,keepaspectratio]{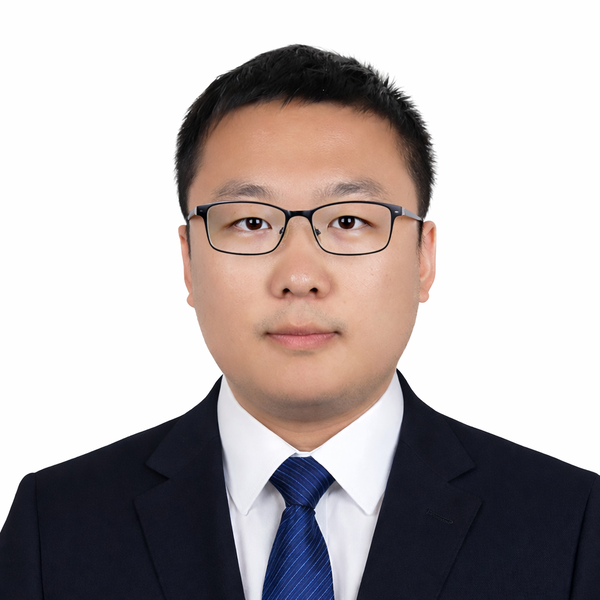}}]{Miles Q. Li}

Miles Q. Li received his Ph.D. in Computer Science from McGill University, Canada, and his M.S. in Computer Software and Theory and B.S. in Physics from Peking University, China. He is currently a Postdoctoral Researcher at McGill University. His research focuses on the security, safety, and trustworthiness of AI systems, including large language model vulnerability analysis, AI agent safety evaluation, adversarial robustness, multi-agent systems, and deep learning for cybersecurity. He has authored multiple peer-reviewed publications in leading journals and conferences in these areas.

\end{IEEEbiography}

\begin{IEEEbiography}[{\includegraphics[width=0.9in,height=1.1in,clip,keepaspectratio]{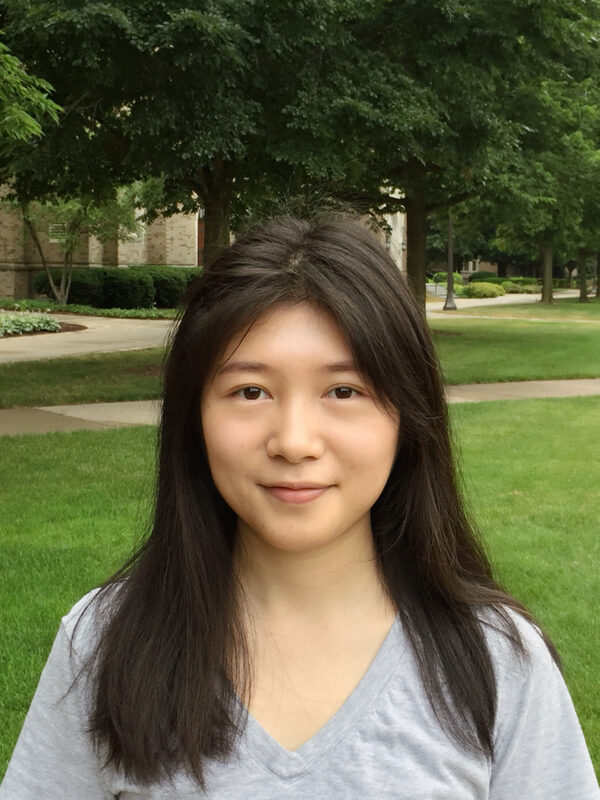}}]{Bingyu Shen}
Bingyu Shen received the B.Eng. degree in information engineering and the M.Eng. degree in communication and information systems from Wuhan University of Technology, Wuhan, China, in 2011 and 2014, respectively, and the Ph.D. degree in computer science and engineering from the University of Notre Dame, Notre Dame, IN, USA, in 2021. She is currently a Senior Machine Learning Engineer at Pinterest. Her research interests include deep learning, machine learning, and large-scale intelligent systems.

\end{IEEEbiography}

\begin{IEEEbiography}[{\includegraphics[width=0.9in,height=1.1in,clip,keepaspectratio]{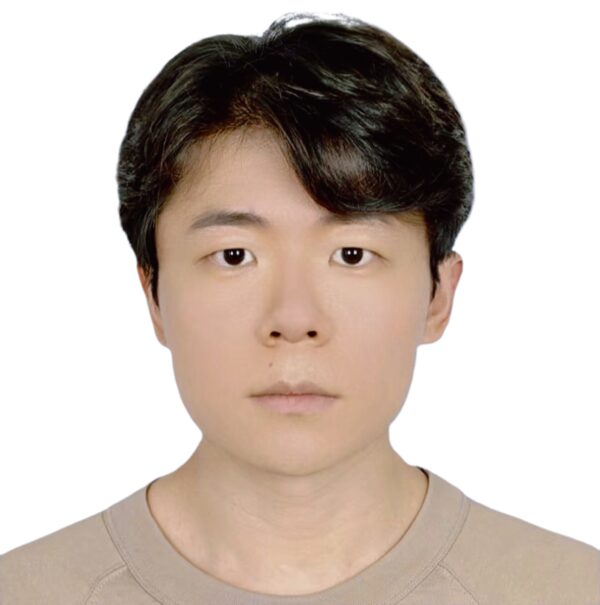}}]{Ruiyang Qin}
Ruiyang Qin (Member, IEEE) received the B.S. and M.S. degrees in computer science from Georgia Institute of Technology in 2020 and 2021, and the Ph.D. degree in computer science and engineering from the University of Notre Dame in 2025. He is currently an Assistant Professor of ECE at Villanova University. His research interests include edge computing, personalized AI, and in-memory computing.
\end{IEEEbiography}

\begin{IEEEbiography}[{\includegraphics[width=0.9in,height=1.1in,clip,keepaspectratio]{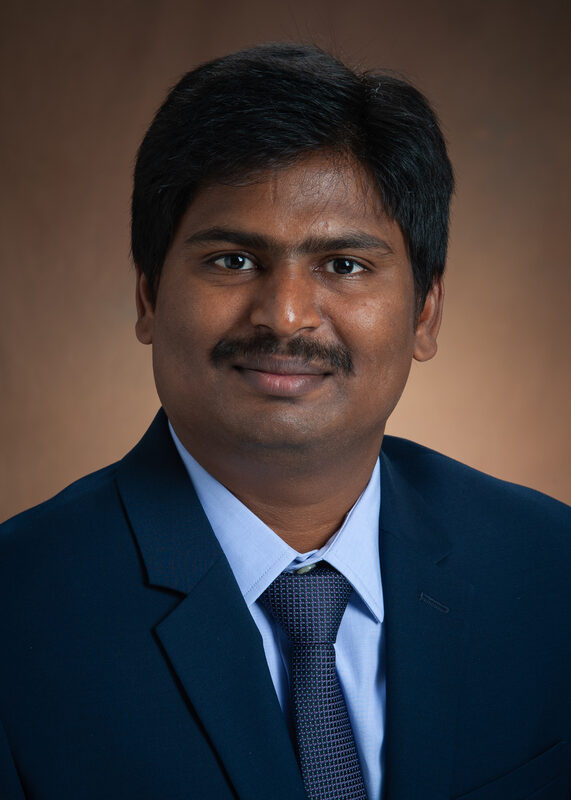}}]{Umamaheswara Rao Tida}
Umamaheswara Rao Tida is an Assistant Professor in the Department of Electrical and Computer Engineering at North Dakota State University. He received the Ph.D. degree in Electrical Engineering from the University of Notre Dame in 2019. His research interests include monolithic 3-D integrated systems, machine learning for IC design, and application-specific machine learning frameworks.
\end{IEEEbiography}

\begin{IEEEbiography}[{\includegraphics[width=0.9in,height=1.1in,clip,keepaspectratio]{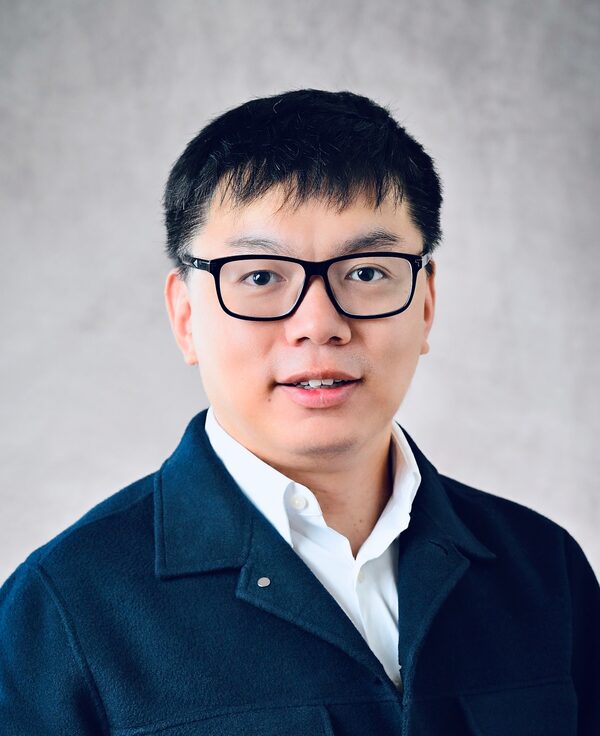}}]{Boyang Li}

Boyang Li (Member, IEEE) is an Assistant Professor in the Department of Computer Science at Kean University. Prior to joining Kean, he worked with Meta (formerly Facebook) as an AI track Research Scientist. He received the Ph.D. degree in Computer Science and Engineering from the University of Notre Dame (US). He received the M.S. degree in Nanoelectronics from the University of Southampton (UK). He received the B.S. degree in Electrical Science and Technology from Xi'an University of Posts and Telecommunications (China). His first-author paper on PoDL received the Best Student Paper Award from the IEEE Biometrics Council (2019, CVPR workshop). His research interests include AI Safety and Blockchain Novel Consensus. 
\end{IEEEbiography}

\end{document}